\documentclass[10pt,twocolumn,letterpaper]{article}
\usepackage{cvpr}

\usepackage[dvipsnames]{xcolor}

\usepackage{float}
\usepackage{times}
\usepackage{epsfig}
\usepackage{graphicx}
\usepackage{amssymb}
\usepackage{amsmath}
\usepackage{mathtools}
\usepackage{booktabs}
\usepackage{subcaption}
\usepackage{comment}
\usepackage{color,colortbl}
\usepackage{multirow}
\usepackage{hhline}
\usepackage{import}
\usepackage{xargs}
\usepackage{bbm}
\usepackage{textcomp}
\usepackage[accsupp]{axessibility}

\definecolor{cvprblue}{rgb}{0.21,0.49,0.74}
\usepackage[pagebackref,breaklinks,colorlinks,citecolor=cvprblue]{hyperref}

\def\paperID{13219}
\def\confName{CVPR}
\def\confYear{2024}

\title{LaneCPP: Continuous 3D Lane Detection using Physical Priors}

\author{Maximilian Pittner$^{1,\,2}$, Joel Janai$^{1}$, Alexandru P. Condurache$^{1,\,2}$ \\ 
	$^{1}$Bosch Mobility Solutions, Robert Bosch GmbH \\
	$^{2}$Institute of Signal Processing, University of L\"ubeck \\
	{\tt\small \{Maximilian.Pittner, Joel.Janai, AlexandruPaul.Condurache\}@de.bosch.com}
}

\def\maketitlesupplementaryarxiv
{
	\newpage
	\twocolumn[
	\centering
		\Large
		\textbf{Supplementary Material} \\
	\vspace{1.0em}
	] 
}

\newcommand{\figref}[1]{Fig.~\ref{#1}}

\newcommand{\tabref}[1]{Table~\ref{#1}}

\DeclareMathOperator*{\argmin}{argmin~}

\makeatletter
\DeclareRobustCommand\onedot{\futurelet\@let@token\@onedot}
\def\@onedot{\ifx\@let@token.\else.\null\fi\xspace}

\makeatother

\definecolor{darkgreen}{rgb}{0,0.7,0}
\definecolor{Gray}{gray}{0.92}

\begin{document}
\maketitle
\begin{abstract}
	Monocular 3D lane detection has become a fundamental problem in the context of autonomous driving, which comprises the tasks of finding the road surface and locating lane markings. One major challenge lies in a flexible but robust line representation capable of modeling complex lane structures, while still avoiding unpredictable behavior. While previous methods rely on fully data-driven approaches, we instead introduce a novel approach LaneCPP that uses a continuous 3D lane detection model leveraging physical prior knowledge about the lane structure and road geometry. While our sophisticated lane model is capable of modeling complex road structures, it also shows robust behavior since physical constraints are incorporated by means of a regularization scheme that can be analytically applied to our parametric representation. Moreover, we incorporate prior knowledge about the road geometry into the 3D feature space by modeling geometry-aware spatial features, guiding the network to learn an internal road surface representation. In our experiments, we show the benefits of our contributions and prove the meaningfulness of using priors to make 3D lane detection more robust. The results show that LaneCPP achieves state-of-the-art performance in terms of F-Score and geometric errors.
\end{abstract}    
\section{Introduction}
\textit{\label{sec:introduction}}
Robust and precise lane detection systems build one of the most essential components in the perception stack of autonomous vehicles. 
While some approaches utilize LiDAR sensors or multi-sensor setups, the application of monocular cameras has become more popular due to their lower cost and the high-resolution visual representation that provides valuable information to detect lane markings.

In the past, lane detection was mainly treated as a 2D detection task. Deep learning based methods achieved good results by treating the problem as a segmentation task in pixel space \cite{vpgnet,scnn,ghafoorian2018gan,lanenet,lightweightld,pizzati2019lane,zou2019robust}, 
used to classify and regress lanes using anchor-based \cite{linecnn,laneatt} representations, or as key-points on a grid structure  \cite{huval2015empirical,pinet,qu2021fololane,wang2022ganet}. 
However, due to the lack of depth information, these 2D representations fail to model lane markings and road geometry in 3D space, which forms an important prerequisite for later functionalities like trajectory planning. 
Consequently, approaches for monocular 3D lane detection were introduced, which adapted lane representations for the 3D domain by modeling vertical anchors \cite{3dlanenet,genlanenet} or local segments on a grid \cite{3dlanenetplus} in a Birds-Eye-View (BEV) oriented 3D-frame. 

A crucial topic for the application of lane detection algorithms in autonomous systems is safety, which requires predictable and robust behavior in any traffic situation. 
One risk of learning-based methods is the tendency to show unpredictable behavior in cases of rarely observed scenarios. 
Since obtaining large amounts of data with high-quality annotations is cumbersome and expensive, publicly available 3D datasets are limited in size and accuracy. Hence, they do not reflect the variability of real-world scenarios sufficiently. This makes learning-based models prone to overfitting, and eventually, diminishes predictability.

One common way to deal with such problems is the integration of prior knowledge. Physics provides us a profound understanding of the 3D world, allowing us to make valid assumptions about the lane structure and road surface geometry. 
Therefore, we introduce physically motivated priors into the lane detection objective to cope with the limited data problem and achieve robust and predictable behavior. 

There are certain geometric properties that should generally hold for detected lane lines. 
For instance, we know that most lines progress parallel to each other, reside on a smooth surface and should not exceed certain thresholds in terms of curvature and slope. However, integrating such assumptions into prevailing discrete representations is not straight forward as strong simplifications are necessary. In contrast, continuous 3D lane representations directly provide parametric curves using polynomials \cite{liu2022learning,bai2023curveformer} or more sophisticated B-Splines \cite{pittner20233d}. These allow for analytical computations on the curve function, which enables the integration of such priors into the lane representation. By modeling these priors explicitly instead of learning them from data, the model can focus its full capacity on learning richer features for the lane detection task.

We can further use physical knowledge about the road geometry to support the model in learning an internal transformation from image features to 3D space. 
While methods based on Inverse Perspective Mapping (IPM) \cite{3dlanenet,3dlanenetplus,genlanenet,liu2022learning,li2022reconstruct,pittner20233d} make false flat-ground assumptions, learning based transformations \cite{chen2022persformer,bai2023curveformer,wang2023bev} completely ignore road properties. 
In contrast, integrating prior knowledge about the road surface allows us to model 3D features geometry-aware and helps the network to focus on the 3D region of interest. 

Thus, we propose a novel 3D lane detection approach named LaneCPP that leverages valuable prior knowledge to achieve accurate and robust perception behavior. It introduces a new sophisticated continuous curve representation, which enables us to incorporate physical priors. 
In addition, we present a spatial transformation component for learning a physically inspired mapping from 2D image to 3D space providing meaningful spatial features. 

\noindent Our main contributions can be summarized as follows:
\begin{itemize}
	\item We propose a novel architecture for 3D lane detection from monocular images using a more sophisticated flexible parametric spline-based lane representation.
	\item We present a way to incorporate priors about lane structure and geometry into our continuous representation. 
	\item We introduce a new way to use prior knowledge about the road surface geometry for learning spatial features. 
	\item We demonstrate the benefits of our contributions in several ablation studies. 
	\item We show state-of-the-art performance of our model. 
\end{itemize}
\begin{figure*}[t]
	\def\svgwidth{.99\linewidth}
		\centering
		\includegraphics[width=0.99\linewidth]{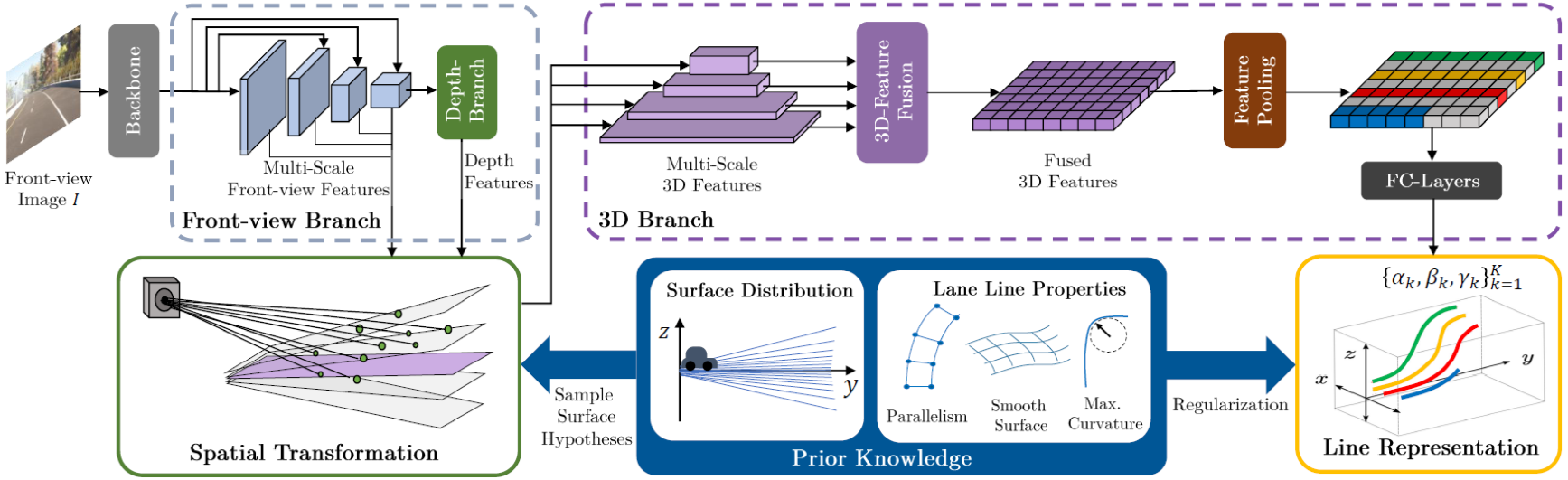}
	\caption{Our approach: First, front-view image $I$ is propagated through the backbone extracting multi-scale feature maps. These are transformed to 3D using our spatial transformation and then fused to obtain a single 3D feature map. Feature pooling is applied to obtain features for each line proposal that are propagated through fully connected layers to obtain the parameters for our line representation. Finally, prior knowledge is exploited to regularize the lane representation and to produce surface hypotheses for the spatial transformation.}	\label{fig:overview}
\end{figure*}

\section{Related work}
\label{sec:relatedwork}
\textbf{Different Lane Representations}. 
An important design choice in deep learning based lane detection is the representation that the network uses to model lane line geometry, which can be categorized as follows:
\textit{1) Pixel-wise} representations, which formulate lane detection as a segmentation problem, were used mainly in 2D methods \cite{vpgnet,scnn,ghafoorian2018gan,lanenet,lightweightld,pizzati2019lane,zou2019robust,zheng2021resa} and were adopted in 3D by SALAD \cite{once3dlanes} combining line segmentation with depth-prediction. These representations come with high computational load since a large amount of parameters is required. 
\textit{2) Grid-based} approaches divide the space into cells and model lanes using local segments \cite{huval2015empirical} or key-points \cite{pinet,qu2021fololane,wang2022ganet}. 3D-LaneNet+ \cite{3dlanenetplus} suggests to use local line-segments and BEV-LaneDet \cite{wang2023bev} defines key-points on a BEV grid representation. Both depend on the grid resolution and require costly post-processing to obtain lines. \textit{3) Anchor-based} representations \cite{linecnn,laneatt,sgld,zheng2022clrnet} model lines as straight anchors with positional offsets at predefined locations. They are widely used in 3D detection approaches including 3D-LaneNet \cite{3dlanenet} and Gen-LaneNet \cite{genlanenet}, which use vertical anchors in the top-view, and Anchor3DLane \cite{huang2023anchor3dlane}, introducing anchor projection with iterative regression. 
Similar to grid-based representations, it requires subsequent curve-fitting to obtain smooth lines. \quad
\noindent \textit{4) Continuous curve} representations \cite{wang2020poly,polylanenet,lstr,lu2021super,curvemodeling} instead directly model smooth curves without requiring costly post-processing. While CLGO \cite{liu2022learning} and CurveFormer \cite{bai2023curveformer} use simple polynomials, \mbox{3D-SpLineNet} \cite{pittner20233d} proposes B-Splines \cite{deboor197250}. Since B-Splines offer local control over curve segments, they are compatible to model complex shapes with low-degree basis functions, while polynomials and B\'ezier curves show global dependence and thus require higher degrees causing expensive computation. Although \mbox{3D-SpLineNet} achieves superior detection performance on synthetic data, it unfortunately lacks flexibility as the curve formulation is limited to monotonically progressing lanes, making it hardly applicable to real-world data. To resolve this issue, we propose a more flexible representation based on actual 3D B-Splines. In contrast to discrete grids and anchors, continuous representation even allow us to integrate prior knowledge in an analytical manner.

\textbf{Geometry Priors}. 
Several approaches suggest to incorporate prior knowledge into learning-based methods, e.g. by integrating invariance into the model architecture \cite{rath2020invariant,rath2022improving} or task-specific transformations as for trajectory planning \cite{eqmotion,wang2023eqdrive,hagedorn2024pioneering}. 
In the field of lane detection, line parallelism has been formulated as a hard constraint to resolve depth ambiguity and determine camera parameters \cite{nieto2008robust,xiong2018}. Deep declarative networks \cite{gould2021deep} offer a general framework to incorporate arbitrary properties as constraints, by solving a constrained optimization problem in the forward pass. 
While such methods are appropriate when hard constraints must be enforced, our goal is rather to guide the network in learning typical geometric lane properties by formulating soft constraints in a regularization objective. Such a regularization only affects training and does not require resolving an optimization problem in the forward pass, and thus, comes without additional computational cost during inference. Following this paradigm, SGNet \cite{sgld,lu2021super} proposes to penalize the deviation of lateral distance from a constant lane width in the IPM warped top-view, but ignores that the property does not hold for lines deviating from the ground plane. 
GP \cite{li2022reconstruct} presents a parallelism loss that enforces constant distance between nearest neighbors locally, which depends on the number of anchor points. In contrast, our method presents a way to learn parallelism globally and independent of resolutions of discrete lane representations. We propose an elegant way to learn parallelism as well as other geometry priors using analytical formulations of tangents and normals, which are well-defined on our continuous spline representation.

\textbf{Leveraging 3D Features}. 
An important model component consists in the extraction of 3D features, encoding valuable information to detect lanes along the road surface. While some works predict 3D lanes directly from the front-view, e.g. by utilizing pixel-wise depth estimation \cite{once3dlanes} or 3D anchor-projection mechanisms \cite{huang2023anchor3dlane}, prevalent methods employ an intermediate 3D or BEV feature representation with an internal transformation from the front-view to the 3D space. 
3D-LaneNet \cite{3dlanenet} proposes to utilize IPM \cite{ipm} to project front-view features to a flat road plane due to the spatial correlation between the warped top-view image and 3D lane geometry and was adopted in several other works \cite{3dlanenetplus,genlanenet,liu2022learning,li2022reconstruct,pittner20233d}. 
However, IPM causes visual distortions in the top-view representation when the flat road assumption is violated. 
In related fields like BEV semantic segmentation, BEV transformations are learned via Multi-Layer-Perceptrons (MLPs) \cite{pan2020cross,li2022hdmapnet}, depth prediction \cite{roddick2019oft,roddick2020pyroccnet,philion2020lift} or transformer-based attention mechanisms \cite{saha2022translating,li2022bevformer,peng2023bevsegformer}. In 3D lane detection, PersFormer \cite{chen2022persformer} utilizes attention between front- and top-view, CurveFormer \cite{bai2023curveformer} introduces dynamic 3D anchors that model queries as parametric curves and BEV-LaneDet \cite{wang2023bev} uses MLPs for the spatial transformations. 
However, these learned transformations do not necessarily provide a 3D feature representation since they are not guided by valuable priors about the road surface geometry, which potentially results in unforeseen behavior for out-of-distribution data. 
Our approach instead aims for carefully modeling a geometry-aware feature space using a depth classification method inspired by \cite{philion2020lift} that exploits knowledge about the distribution of the road surface. 
\section{Methodology} 
\label{sec:methodology}
The following section describes our 3D lane detection approach. An overview of the overall architecture is described and illustrated in \figref{fig:overview}. The main focus lies on our continuous 3D lane line representation, our regularization mechanism using physical priors and our prior-based spatial transformation module, which we explain in the following. 

\begin{figure}[tb]
		\centering
		\includegraphics[width=0.99\linewidth]{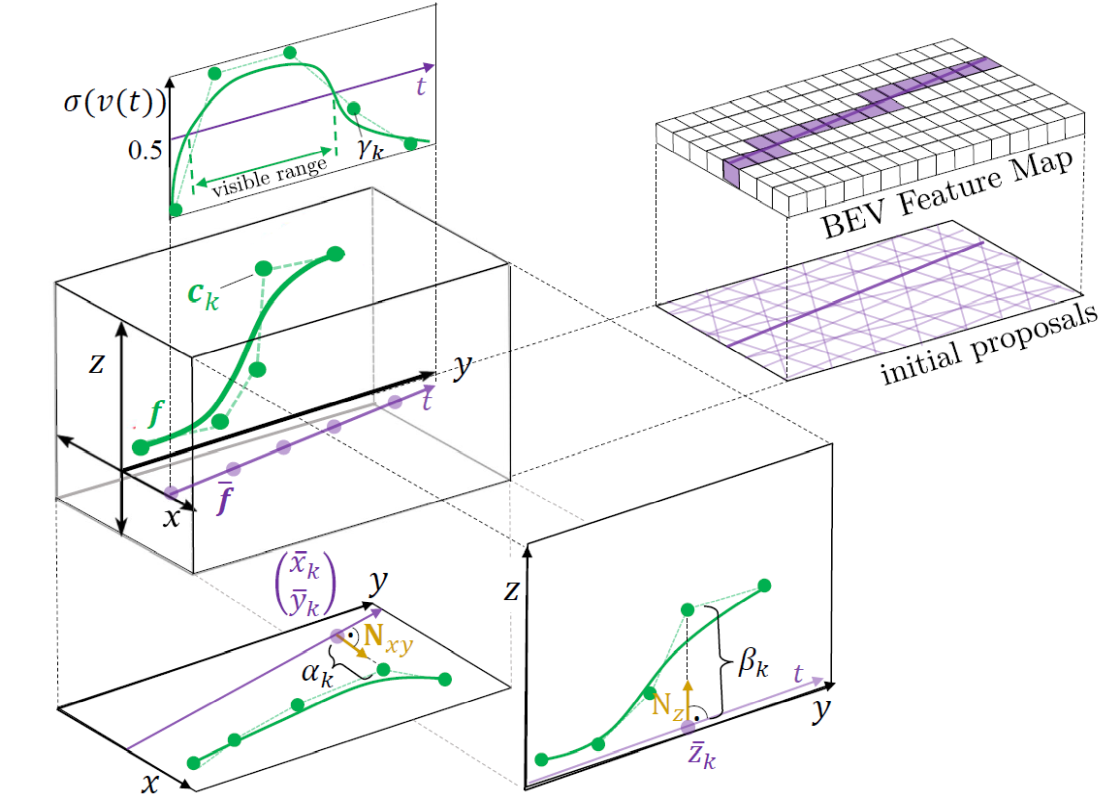}
	\caption{Our 3D lane line representation: For each proposal $\boldsymbol{\bar{f}}$ (\textcolor{violet}{purple lines}), line geometry is described by 3D B-Splines with control points $\boldsymbol{c}_k$ (\textcolor{darkgreen}{green dots}). Each control point is determined by the offsets $\alpha_k, \, \beta_k$ from the control points of the initial proposal in normal direction (\textcolor{orange}{orange vectors}). Additionally, visibility $v(t)$ is modeled by splines with 1D control points $\gamma_k$.}
	\label{fig:rep}
\end{figure}

\subsection{Lane line representation}
\label{subsec:rep}
Inspired by prior work in 3D lane detection \cite{pittner20233d}, we leverage the benefits of continuous representations and employ a parametric model based on B-Splines. 
However, modeling only lateral ($x$-) and vertical ($z$-) components with spline-based functions (as done in previous approaches) is limited to lanes that merely progress along the longitudinal ($y$-) direction. 
Instead, we propose the first full 3D lane line representation modeling each component ($x$, $y$, $z$) such that we obtain 
\begin{align}
\boldsymbol{f}(t) = 
\begin{pmatrix}
x(t)   \\
y(t)	\\
z(t)
\end{pmatrix}
= \sum_{k=1}^{K} \, \boldsymbol{c}_k \cdot B_{k,d}(t) \, 
\end{align}
with curve argument $t \in [0,\,1]$ and $K$ control points $\boldsymbol{c}_k = {\big(x_k, \, y_k, \,z_k\big)}^T$. Each control point  $\boldsymbol{c}_k$ weights the respective basis function $B_{k,d}(t)$ (recursive polynomials of degree $d$) controlling the curve shape. 

Due to the ambiguity of curves using 3D B-Splines (the same spline curve can be described by different configurations of its control points), regressing all three dimensions per control point results in strong overfitting during training. 
We resolve this issue by limiting the degrees of freedom per control point to two and constraining the control points deflection to one direction in the $x$-$y$-plane and one direction in the $y$-$z$-plane as illustrated in \figref{fig:rep}. More precisely, the degrees of freedom per control point are specified by the directions of the normals $\mathbf{N}_{xy}$ and $\mathrm{N}_{z}$ of an initial curve proposal $\boldsymbol{\bar{f}}$ with control points $\boldsymbol{\bar{c}}_k = {\big(\bar{x}_k, \, \bar{y}_k, \,\bar{z}_k\big)}^T$. The control points are then defined as
\begin{align}
\boldsymbol{c}_k =
\begin{pmatrix*}[r]
 x_k \\
 y_k \\
 z_k \\
\end{pmatrix*} 
= 
\begin{pmatrix*}[r]
\bar{x}_k + \mathrm{N}_{x} \cdot \alpha_k \\
\bar{y}_k + \mathrm{N}_{y} \cdot \alpha_k \\
\bar{z}_k + \mathrm{N}_{z} \cdot \beta_k \\
\end{pmatrix*} \,, 
\label{eq:cpconstrained}
\end{align}
where $\mathrm{N}_{x}$, $\mathrm{N}_{y}$ describe the $x$- and $y$-component of the normal vector $\mathbf{N}_{xy}$ in the $x$-$y$-plane.
As shown in \cref{eq:cpconstrained} and illustrated in \figref{fig:rep}, modeling splines as deflections in normal direction of its underlying initial line proposal only requires two parameters $\alpha_k, \beta_k$ per control point to describe the 3D shape. We use a wide variety of orientations for the initial proposals $\boldsymbol{\bar{f}}$ (see \figref{fig:rep}), which allows us to detect any kind of lines with this formulation. More details about the initial proposals are provided in the supplementary.

While \cite{pittner20233d} models the curve range using start- and end-points that are learned by means of regression, we instead propose to model visibility\footnote{For the concept of visibility, we follow the prevailing definition from the literature \cite{genlanenet,chen2022persformer}.} using a continuous representation $v(t)$ and treat the visibility estimation as a classification problem. We obtain probability values applying sigmoid activation and consider $\sigma\big(v(t)\big) > 0.5$ the visible range. While in theory any kind of function can be utilized, we found that B-Splines with the same configuration as $\boldsymbol{f}(t)$ are well-suited and introduce spline control points $\gamma_k$ defining the shape of $v(t)$. 

Eventually, binary cross-entropy is used as a classification loss to learn visibility 
\begin{align}
	\mathcal{L}_{vis} =& - \frac{1}{|\mathcal{P}_{GT}|} \sum_{\boldsymbol{p} \in \mathcal{P}_{GT}} \hat{v}_{\boldsymbol{p}} \cdot \log\big(\sigma \big(v(t_{\boldsymbol{p}})\big)\big) + \\ 
	&(1- \hat{v}_{\boldsymbol{p}}) \cdot \log\big(1 - \sigma \big(v(t_{\boldsymbol{p}})\big)\big) \, ,
\end{align}
where $\mathcal{P}_{GT}$ denotes the ground truth set of points, \mbox{$\hat{v}_{\boldsymbol{p}} \in \{0,\,1\}$} the ground truth visibility for point $\boldsymbol{p}$. $t_{\boldsymbol{p}}$ represents the respective curve argument obtained by orthogonal projection of $\boldsymbol{p}$ onto the underlying line proposal.

\begin{figure}[tb]
		\centering
		\includegraphics[width=0.95\linewidth]{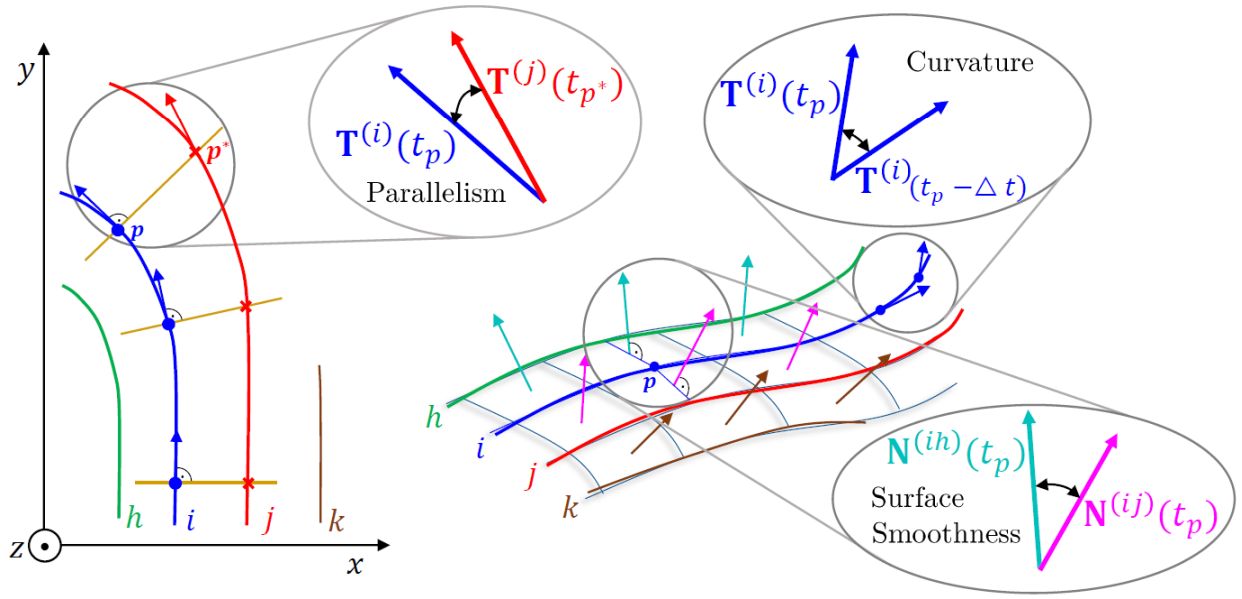}
	\caption{Illustration of different priors expressed by line tangents and surface normals.}
	\label{fig:priors}
\end{figure}

\subsection{Regularization using physical priors}

In this section, we describe our regularization method to integrate prior knowledge about lane structure and surface geometry into our parametric line representation (see \figref{fig:priors}).

\textbf{Line parallelism.} 
In order to reinforce parallel lines, the tangents at point pairs located in opposite normal direction on neighboring lines must be similar (see \figref{fig:priors} left). We realize this by penalizing the cosine distance of the unit tangents $\mathbf{T}(t)$ on neighboring lines $i$ and $j$ for normal point pairs. 
More precisely, for each point ${\boldsymbol{p}} \in \mathcal{P}^{(i)}$ on line $i$ we select the normal pair point ${\boldsymbol{p}}^*$ on neighbor line $j$ that minimizes the distance to the normal plane, which is defined by the plane equation $\mathbf{T}^{(i)}(t)^T \cdot \big((x,y,z)^T - \boldsymbol{f}^{(i)}(t)\big) = 0$. In \figref{fig:priors} the normal planes are visualized in a 2D top-view as lines (orange) for simplicity. 
Hence the respective curve argument $t_{{\boldsymbol{p}}^*}$ for point ${\boldsymbol{p}}^*$ on line $j$ is given as
\begin{align}
	t_{{\boldsymbol{p}}^*} = \argmin_{{{\boldsymbol{p}}' \in \mathcal{P}^{(j)}}} \mathbf{T}^{(i)}(t_{\boldsymbol{p}})^T \cdot \big(\boldsymbol{f}^{(j)}({t_{{\boldsymbol{p}}'}}) - \boldsymbol{f}^{(i)}(t_{\boldsymbol{p}})\big)\,,
	\label{eq:normalpartner}
\end{align}
where $\mathcal{P}^{(j)}$ denotes the points on line $j$. 
While in theory \cref{eq:normalpartner} can be solved analytically, the simpler way is to sample the set of points $\mathcal{P}^{(j)}$ instead. (Note that our continuous representation allows us to choose high sampling rates without losing precision as no interpolation is required.)

With the normal point pairs, we define the parallelism loss for a neighbor line pair based on the cosine distance of their tangents as
\begin{align}
	\mathcal{L}^{(ij)}_{par} = \frac{\mathbbm{1}^{(ij)}_{{\boldsymbol{p}}}}{|\mathcal{P}^{(i)}|} \cdot \sum_{{\boldsymbol{p}} \in \mathcal{P}^{(i)}} 1 - \big(\mathbf{T}^{(i)}(t_{\boldsymbol{p}})\big)^T \cdot \mathbf{T}^{(j)}(t_{{\boldsymbol{p}}^*})\, . 
\end{align}
 Since the criterion of line parallelism should not hold for all normal point pairs of neighboring lines (e.g. merging or splitting lines), $\mathbbm{1}^{(ij)}_{{\boldsymbol{p}}} \in \{0,\,1\}$ represents the indicator function determining whether the parallelism loss is applied to the point pair. More precisely, the function ensures that only the overlapping range of neighboring lines is taken into account. Furthermore, it determines whether the line pair should be considered as a parallel pair based on the standard deviation of euclidean distances between normal point pairs, i.e. high deviations indicate that the line pair might belong to a merge or split structure. 
 In our experiments, we achieve state-of-the-art performance on test sets containing merges and splits, proving that our model is also capable of learning non-parallel line pairs using this indicator function.

\textbf{Surface smoothness.} 
Since the lines reside on a smooth road, the surface normals of neighboring lanes should be similar. Analogously to $\mathcal{L}_{par}$, we express this with the cosine distance between surface normals $\mathbf{N}^{(ih)}$ and $\mathbf{N}^{(ij)}$ as
\begin{align}
\mathcal{L}^{(i)}_{sm} = \frac{\mathbbm{1}^{(hij)}_{{\boldsymbol{p}}}}{|\mathcal{P}^{(i)}|} \cdot \sum_{{\boldsymbol{p}} \in \mathcal{P}^{(i)}} 1 - \big(\mathbf{N}^{(ih)}(t_{\boldsymbol{p}})\big)^T \cdot \mathbf{N}^{(ij)}({t_{\boldsymbol{p}}})\, , 
\end{align}
with indicator function $\mathbbm{1}^{(hij)}_{{\boldsymbol{p}}}$. 
The surface normal between line $i$ and left neighbor line $h$  at point ${\boldsymbol{p}}$ can be expressed as the cross product of the tangent on line $i$ and the normalized connection vector between lines $i$ and $h$, hence $\mathbf{N}^{(ih)}(t_{\boldsymbol{p}}) = \mathbf{T}^{(i)}(t_{\boldsymbol{p}}) \times \frac{\boldsymbol{f}^{(h)}(t_{{\boldsymbol{p}}^*})-\boldsymbol{f}^{(i)}(t_{\boldsymbol{p}})}{|| \boldsymbol{f}^{(h)}(t_{{\boldsymbol{p}}^*})-\boldsymbol{f}^{(i)}(t_{\boldsymbol{p}}) ||}$. 
For the normal between line $i$ and right neighbor $j$ the sign is flipped to obtain upwards pointing normal vectors.

\textbf{Curvature.} 
We determine lane curvature by computing the second order derivatives as the difference of tangents at consecutive points divided by their euclidean distance as $\mathbf{T}'(t_{\boldsymbol{p}})=\frac{\mathbf{T}(t_{\boldsymbol{p}})-\mathbf{T}(t_{\boldsymbol{p}} - \Delta t)}{|| \boldsymbol{f}(t_{\boldsymbol{p}}) - \boldsymbol{f}(t_{\boldsymbol{p}} - \Delta t) ||}$. 
The maximum curvature in $x$-$y$-plane (inverse curve radius) and in $z$ (rate of slope change) have very different value ranges and are therefore restricted by different limits. 
Hence, we define the two thresholds $\kappa_{xy}$  and $\kappa_z$ and formulate the curvature loss on line $i$ as 
\begin{align}
	\mathcal{L}^{(i)}_{curv} =& \frac{1}{|\mathcal{P}^{(i)}|} \cdot \sum_{{\boldsymbol{p}} \in \mathcal{P}^{(i)}} \max \big( \mathrm{T}'^{(i)}_{xy}(t_{\boldsymbol{p}}), \, \kappa_{xy} \big)  \\
	 & + \max \big( \mathrm{T}'^{(i)}_{z}(t_{\boldsymbol{p}}), \, \kappa_z \big) \, .
\end{align}

\noindent Finally, the prior regularization loss is given as 
\begin{align}
	\mathcal{L}_{prior} = \sum_{i=1}^{M} \lambda_{sm} \mathcal{L}^{(i)}_{sm} + \lambda_{curv} \, \mathcal{L}^{(i)}_{curv} + \sum_{j=1}^{N} \lambda_{par} \, \mathcal{L}^{(ij)}_{par} \, ,
\end{align}
with individual weights $\lambda_{par},\,\lambda_{sm},\,\lambda_{curv}$. 
Note that all these properties are expressible by means of tangents and normals, which can be computed analytically on our parametric representation in continuous space. Consequently, minimization of the herein introduced prior losses does not depend on numerical approximations as is the case for anchor-, grid- or key-point representations.

\begin{figure}[tb]
		\centering
		\includegraphics[width=0.97\linewidth]{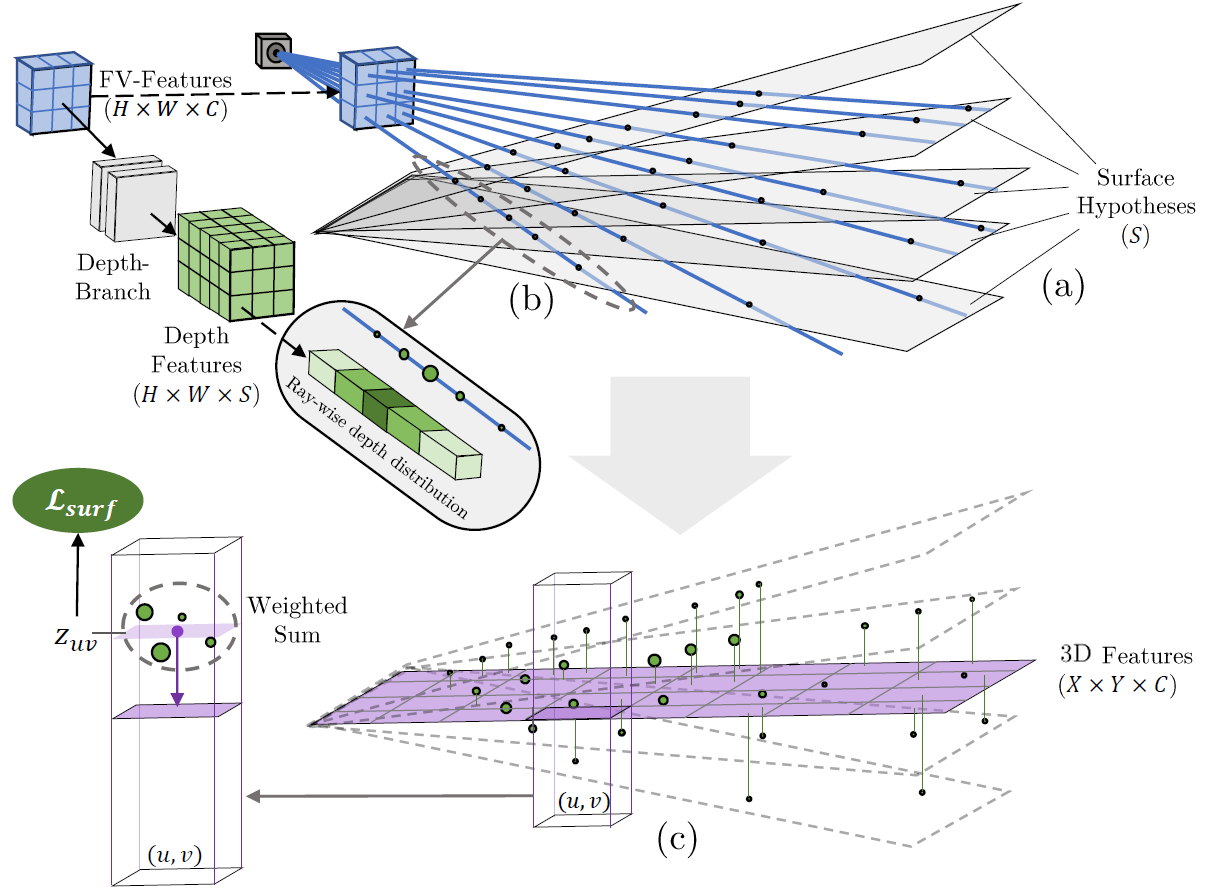}
	\caption{Our proposed spatial transformation module. First, several road surface hypotheses are defined (a) to which front-view features are lifted (b) and weighted according to the predicted depth distribution. Afterwards, point features are aggregated in a weighted manner to obtain the 3D feature map (c).}	\label{fig:spatialtrafo}
\end{figure}

\subsection{Spatial transformation}
In this section, we describe our spatial transformation (shown in \figref{fig:spatialtrafo}) that is leveraging valuable physical knowledge about surface geometry. 
We know that the road surface typically shows small deviations from the ground level ($z=0$) in the near-range and stronger deviations in the far-range. Based on this knowledge, we sample ground surface hypotheses that reflect the distribution of the road surface height profile (\figref{fig:spatialtrafo}a). While in theory different types of surface functions could be utilized as hypotheses, we decide to merely rely on planes, since this facilitates the computation of ray intersections described in the following step.

Next, the multi-scale front-view feature maps extracted by the backbone are lifted to 3D space (\figref{fig:spatialtrafo}b). Our approach is inspired by \cite{philion2020lift}, where front-view features are spreading along rays throughout the space of the road surface. These rays intersect with the surface hypotheses at different depths spanning a frustum-like point cloud in 3D space, where each point is affiliated with a $C$-dimensional feature vector and additionally attached with its height value $z$, hence, each point in the cloud has dimension $(C+1)$. 
The front-view feature map is propagated through a depth branch with a channel-wise softmax applied to obtain a categorical distribution for each ray, resulting in a tensor of size $H \times W \times S$, where $H$, $W$ denote height and width and channel size $S$ the number of surface hypotheses. 

In order to aggregate the information in 3D space, a BEV grid of size $X \times Y$ is defined. Features from points mapping to the same grid cell are weighted by the categorical depth distribution for the respective ray and accumulated in terms of a weighted sum (\figref{fig:spatialtrafo}c). Since the $z$-component of the points is also combined by a weighted sum, the value $z_{uv}$ can be interpreted as the height value of the surface for grid cell $(u,\,v)$. 
We guide the model in learning the real surface and prevent it from learning an arbitrary mapping by introducing a simple grid-based regression loss as 
\begin{align}
	\mathcal{L}_{surf} = \frac{1}{X \cdot Y} \sum_{(u,v) \in X \times Y} \mathbbm{1}_{uv} \cdot {\| z_{uv} - 	\hat{z}_{uv} \|}_1 \, ,
\end{align}
with $\mathbbm{1}_{uv}$ indicating whether surface ground truth $\hat{z}_{uv}$ is available for cell $(u,\,v)$. The height ground truth is obtained by interpolation of the 3D lane annotations at cell locations.

\subsection{Loss functions}\label{subsec:training}
The overall loss used during training is given as the weighted sum of loss components
\begin{align}
\mathcal{L} =& \lambda_{pr} \mathcal{L}_{pr} + \lambda_{cat} \mathcal{L}_{cat} + \lambda_{reg} \mathcal{L}_{reg} + \\ 
&\lambda_{vis} \mathcal{L}_{vis} + \lambda_{prior} \mathcal{L}_{prior} + \lambda_{surf} \mathcal{L}_{surf}  \, .
\end{align}
We use focal loss \cite{lin2017focal} for lane presence $\mathcal{L}_{pr}$ and category classification $\mathcal{L}_{cat}$. For the regression loss $\mathcal{L}_{reg}$, we adapt the formulation of \cite{pittner20233d} to three instead of two dimensions. More details are provided in the supplementary. 
\begin{table}
	\centering
	\resizebox{\columnwidth}{!}{
		\begin{tabular}{lccccc}
			\toprule
			\bf{Priors} & \bf{F1(\%)}$\uparrow$ & \bf{X-near(m)}$\downarrow$ & \bf{X-far(m)}$\downarrow$ & \bf{Z-near(m)}$\downarrow$ & \bf{Z-far(m)}$\downarrow$ \\ 
			\hhline{======}
			\bf{None} 			& $65.0$ 	& $0.316$ 	& $0.384$ 	& $0.106$ 	& $0.153$ 	\\
			\bf{Par.} 			& $66.2$ 	& $0.291$ 	& $0.373$ 	& $0.103$ 	& $0.150$ 	\\
			\bf{Surf.} 			& $65.8$	& $0.320$	& $0.356$	& $0.103$	& $0.144$	\\
			\bf{Curv.}			& $66.7$	& $0.322$	& $0.366$	& $0.105$	& $0.146$	\\
			\rowcolor{Gray} \bf{Comb.} 			& $66.7$	& $0.301$	& $0.359$	& $0.103$	& $0.144$	\\
			\bottomrule
	\end{tabular}}
	\caption{Effect of different prior losses on OpenLane300.} \label{tab:priors}
\end{table}

\begin{table}
	\centering
	\resizebox{\columnwidth}{!}{
		\begin{tabular}{c||cc>{\columncolor{Gray}}ccc}
			\toprule
			\bf{\# Surface Hypotheses} & \bf{1} & \bf{3} & \bf{5} & \bf{15} & \bf{27} \\ 
			\hline
			\bf{F1-Score(\%)}$\uparrow$	& 65.0 & 65.9 & \bf{66.6} & 66.1 & 66.0 \\
			\bottomrule
		\end{tabular}
	}
	\caption{Effect of the surface hypotheses on OpenLane300.} \label{tab:spatialtrafo}
\end{table}

\begin{table}
	\centering
	\resizebox{\columnwidth}{!}{
		\begin{tabular}{ccc||cc}
			\toprule
			\bf{Lane Rep.} & \bf{Prior Reg.} & \bf{Spatial T.} & \bf{F1(\%)}$\uparrow$ & \bf{Gain(\%)} \\ 
			\hhline{=====}
			& & &  												$62.9$  & 	 (baseline)				\\
			\checkmark & & &  									$65.0$ & 			$+2.1$ 				\\
			\checkmark & \checkmark & &  						$66.7$ & 			$+3.8$				\\
			\checkmark & & \checkmark &  						$66.6$ & 			$+3.7$ 				\\
			\checkmark & \checkmark & \checkmark &  			$\mathbf{66.9}$ &	$\mathbf{+4.0}$ 	\\
			\bottomrule
	\end{tabular}}
	\caption{Performance gain for different contributions on OpenLane300 using our novel \textbf{Lane Representation}, \textbf{Prior Regularization} and \textbf{Spatial Transformation} instead of IPM.} \label{tab:benefits}
\end{table}

\begin{figure}[tb]
		\centering
		\includegraphics[width=0.9\linewidth]{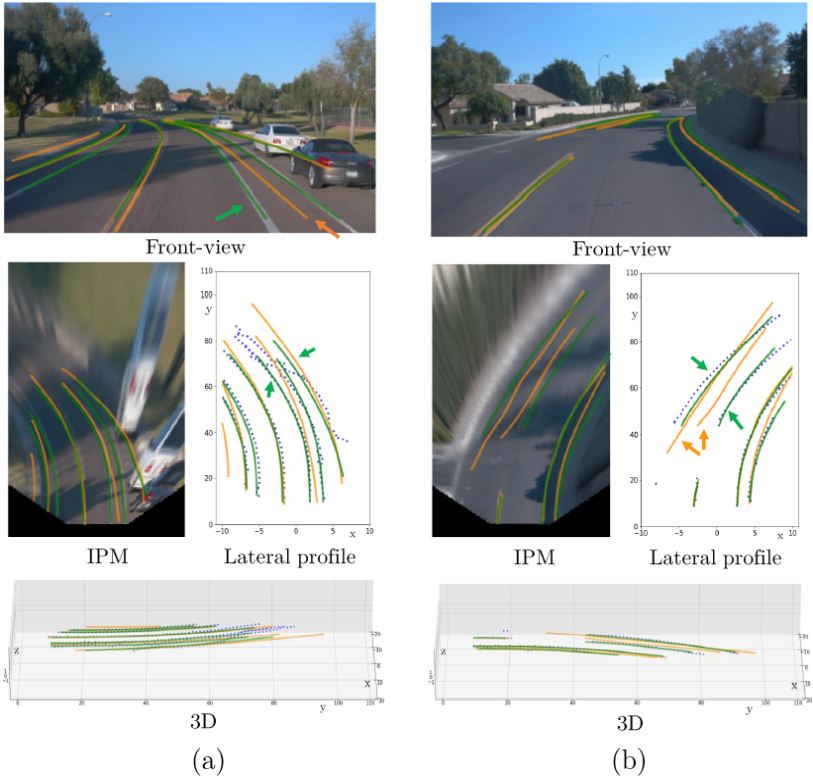}
	\caption{Qualitative comparison of \textcolor{darkgreen}{our model trained with prior regularization} to the same \textcolor{orange}{model without regularization} both trained on OpenLane300 with main differences highlighted by arrows. As a reference \textcolor{blue}{ground truth} lines are visualized dashed.}	\label{fig:parallelism}
\end{figure}

\begin{table*}[tb]
		\centering
		\resizebox{\linewidth}{!}{
			\begin{tabular}{l||ccccc|cccccc}
				\toprule
				\multirow{2}{*}{\bf{Method}} & \multirow{2}{*}{\bf{F1-Score(\%)}$\uparrow$} & \bf{X-error} & \bf{X-error} & \bf{Z-error} & \bf{Z-error} & \multicolumn{6}{c}{\bf{F1-Score(\%) per Scenario} $\uparrow$} \\ 
				&  & \bf{near(m)}$\downarrow$ & \bf{far(m)}$\downarrow$ & \bf{near(m)}$\downarrow$ & \bf{far(m)}$\downarrow$ & \bf{U\&D} & \bf{C} & \bf{EW} & \bf{N} & \bf{I} & \bf{M\&S} \\
				\hhline{=||=====|======}
				3D-LaneNet \cite{3dlanenet} 				& $44.1$ 			& $0.479$ 			& $0.572$				& $0.367$ 			& $0.443$ & $40.8$ 			& $46.5$ 			& $47.5$				& $41.5$ 			& $32.1$ 			& $41.7$ 			\\
				Gen-LaneNet \cite{genlanenet} 				& $32.3$ 			& $0.591$ 			& $0.684$ 				& $0.411$ 			& $0.521$ 
				& $25.4$  			& $33.5$ 			& $28.1$ 				& $18.7$ 			& $21.4$ 			& $31.0$ 			\\
				PersFormer \cite{chen2022persformer} 		& $50.5$ 			& $0.485$ 			& $0.553$ 				& $0.364$ 			& $0.431$ 
				& $42.4$			& $55.6$ 			& $48.6$ 				& $46.6$ 			& $40.0$ 			& $50.7$ 			\\
				PersFormer* \cite{chen2022persformer} 	& $53.1$ 				& $0.361$			& $0.328$ 			&$0.124$ 				& $\underline{0.129}$ 		&$46.8$ 			& $58.7$ 			& $\underline{54.0}$ 				&$48.4$ 			& $41.4$ 			& $52.5$ \\
				CurveFormer \cite{bai2023curveformer} 		& $50.5$ 			& $0.340$ 			& $0.772$ 				& $0.207$ 			& $0.651$ 
				& $45.2$ 			& $56.6$ 			& $49.7$ 				& $49.1$ 			& $42.9$ 			& $45.4$ 			\\
				BEV-LaneDet \cite{wang2023bev}  			& $\underline{58.4}$ 			& $0.309$ 			& $0.659$ 				& $0.244$ 			& $0.631$ 
				& $\underline{48.7}$ 			& $\underline{63.1}$ 			& $53.4$ 				& $\underline{53.4}$ 			& $\underline{50.3}$ 			& $\underline{53.7}$ 			\\
				Anchor3DLane \cite{huang2023anchor3dlane}	& $53.7$ 			& $0.276$ 			& $\underline{0.311}$		 	& $0.107$ 			& $0.138$ 
				& $46.7$			& $57.2$ 			& $52.5$ 				& $47.8$ 			& $45.4$ 			& $51.2$ 			\\
				Anchor3DLane-T \cite{huang2023anchor3dlane}	& $54.3$ 			& $\underline{0.275}$ 			& $\mathbf{0.310}$	& $\underline{0.105}$ 			& $0.135$ 
				& $47.2$			& $58.0$ 			& $52.7$ 				& $48.7$ 			& $45.8$ 			& $51.7$ 			\\
				\rowcolor{Gray} LaneCPP (Ours) & $\mathbf{60.3}$ 	& $\mathbf{0.264}$ & $\mathbf{0.310}$ 		& $\mathbf{0.077}$	& $\mathbf{0.117}$ 
				& $\mathbf{53.6}$ 	& $\mathbf{64.4}$	& $\mathbf{56.7}$	 	& $\mathbf{54.9}$	& $\mathbf{52.0}$ 	& $\mathbf{58.7}$	\\
				\bottomrule
		\end{tabular}}
		\caption{Quantitative comparison on OpenLane \cite{chen2022persformer}. \textbf{Best performance} and \underline{second best} are highlighted. 
			The scenario categories are Up and Down ({U\&D}), Curve ({C}), Extreme Weather ({EW}), Night ({N}), Intersection ({I}), Merge and Split ({M\&S}). 
			{PersFormer*} denotes the latest performance reported on the official code base, Anchor3DLane-T represents the temporal multi-frame method of \cite{huang2023anchor3dlane}.}
		\label{tab:comparison-quant-ol}
\end{table*}

\begin{figure*}
	\centering
	\begin{subfigure}{.197\textwidth}
		\centering
		\includegraphics[width=.99\linewidth]{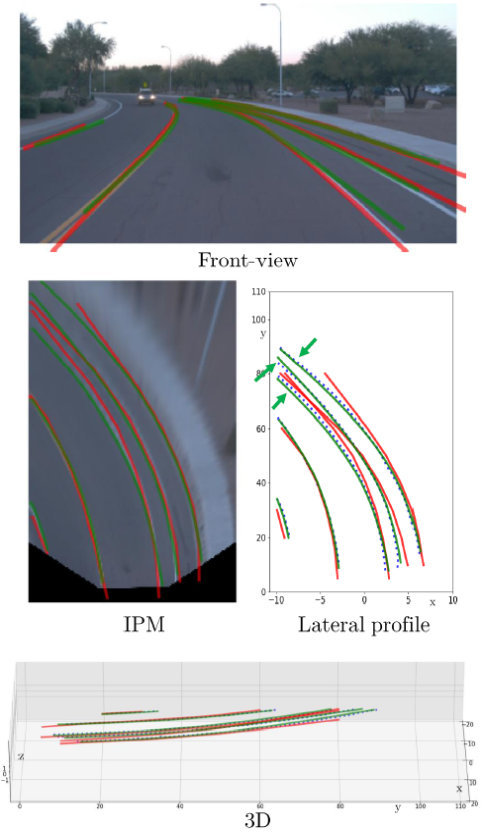}
		\caption{}
		\label{fig:sota-comp-a}
	\end{subfigure}%
	\begin{subfigure}{.197\textwidth}
		\centering
		\includegraphics[width=.99\linewidth]{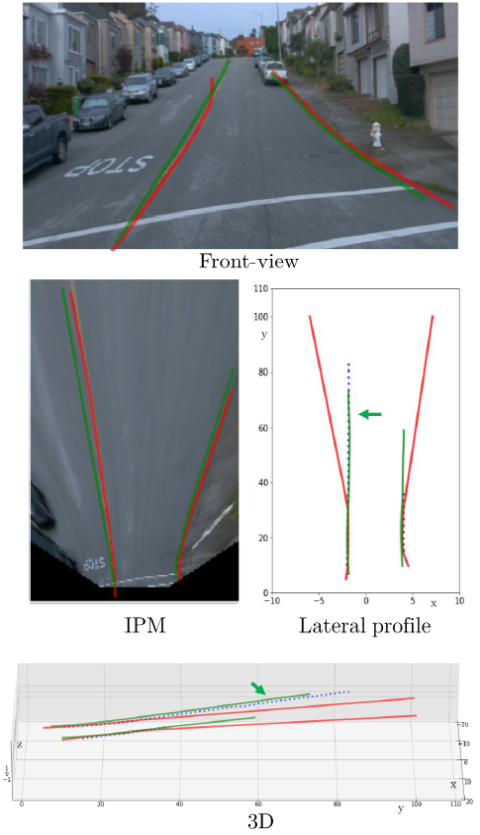}
		\caption{}
		\label{fig:sota-comp-b}
	\end{subfigure}
	\begin{subfigure}{.197\textwidth}
		\centering
		\includegraphics[width=.99\linewidth]{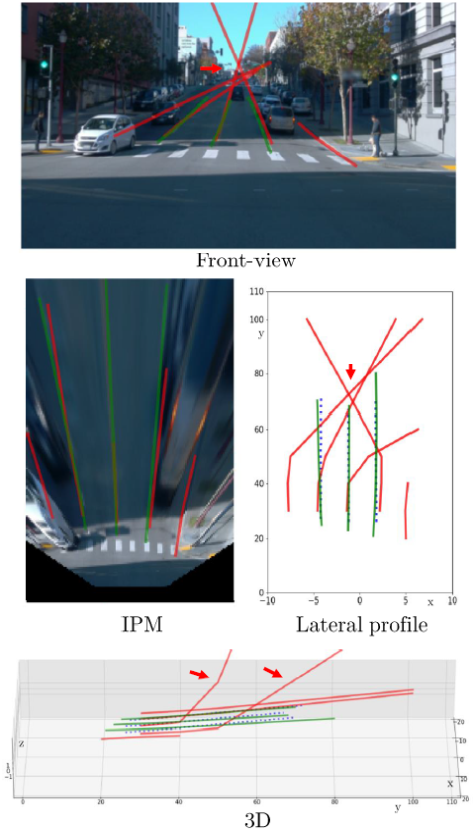}
		\caption{}
		\label{fig:sota-comp-c}
	\end{subfigure}
	\begin{subfigure}{.197\textwidth}
		\centering
		\includegraphics[width=.99\linewidth]{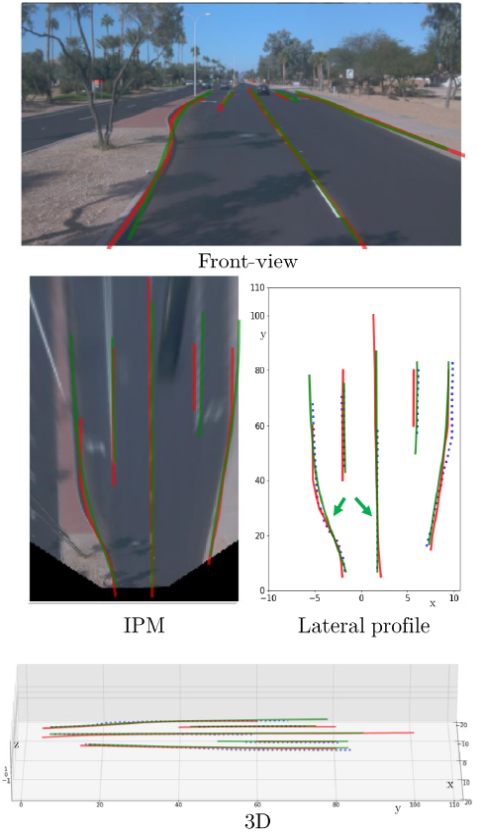}
		\caption{}
		\label{fig:sota-comp-d}
	\end{subfigure}
	\begin{subfigure}{.197\textwidth}
		\centering
		\includegraphics[width=.99\linewidth]{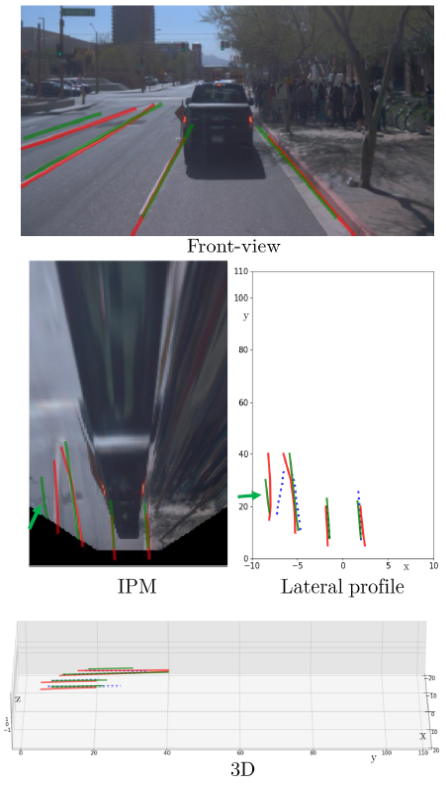}
		\caption{}
		\label{fig:sota-comp-e}
	\end{subfigure}
	\caption{Qualitative comparison on OpenLane. \textcolor{darkgreen}{Our method} is compared to \textcolor{red}{PersFormer*} with \textcolor{blue}{ground truth} visualized as dashed lines.}
	\label{fig:sota-comp}
\end{figure*}

\begin{table*}[tb]
	\centering
	\resizebox{\linewidth}{!}{
		\begin{tabular}{l||ccccc|ccccc}
			\toprule
			\multirow{3}{*}{\bf{Method}} & \multicolumn{5}{c|}{\bf{Balanced Scenes}} & \multicolumn{5}{c}{\bf{Rare Scenes}} \\
			\cline{2-11}
			& \multirow{2}{*}{\bf{F1(\%)}$\uparrow$} & \multicolumn{2}{c}{\bf{X-error (m)} $\downarrow$} & \multicolumn{2}{c|}{\bf{Z-error (m)} $\downarrow$} 
			& \multirow{2}{*}{\bf{F1(\%)}$\uparrow$} & \multicolumn{2}{c}{\bf{\bf{X-error (m)} $\downarrow$}} & \multicolumn{2}{c}{\bf{\bf{Z-error (m)} $\downarrow$}} \\
			& & \bf{near} & \bf{far} & \bf{near} & \bf{far} & & \bf{near} & \bf{far} & \bf{near} & \bf{far} \\
			\hhline{=||=====|=====} 
			3D-LaneNet \cite{3dlanenet} 					& $86.4$  			& $0.068$ 			& $0.477$ 				& $0.015$ 			& $\mathbf{0.202}$ 
			& $72.0$			& $0.166$ 			& $0.855$ 				& $0.039$ 			& $\mathbf{0.521}$ \\
			GP \cite{li2022reconstruct} 					& $91.9$ 			& $0.049$ 			& $0.387$ 				& $\mathbf{0.008}$ 	& $0.213$ 
															& $83.7$ 			& $0.126$ 			& $0.903$ 				& $\underline{0.023}$ 			& $0.625$ \\ 		
			PersFormer \cite{chen2022persformer} 			& $92.9$ 			& $0.054$ 			& $0.356$ 				& $0.01$ 			& $0.234$ 
			& $87.5$ 			& $0.107$ 			& $0.782$ 				& $0.024$ 			& $0.602$ \\
			3D-SpLineNet \cite{pittner20233d}				& $96.3$			& $0.037$			& $0.324$				& $\underline{0.009}$			& $0.213$ 
			& $92.9$			& $0.077$			& $0.699$				& $\mathbf{0.021}$	& $0.562$ \\
			CurveFormer \cite{bai2023curveformer}			& $95.8$ 			& $0.078$ 			& $0.326$ 				& $0.018$ 			& $0.219$ 
			& $95.6$ 			& $0.182$ 			& $0.737$ 				& $0.039$ 			& $0.561$ \\
			BEV-LaneDet \cite{wang2023bev}  				& $96.9$ 			& $\mathbf{0.016}$ 	& $\mathbf{0.242}$ 		& $0.02$ 			& $0.216$ 
			& $\mathbf{97.6}$ 	& $\mathbf{0.031}$ 	& $\mathbf{0.594}$ 		& $0.040$ 			& $0.556$ \\
			Anchor3DLane \cite{huang2023anchor3dlane}		& $95.4$			& $0.045$			& $0.300$				& $0.016$ 			& $0.223$ 
			& $94.4$			& $0.082$			& $0.699$				& $0.030$ 			& $0.580$ \\
			\rowcolor{Gray} LaneCPP (Ours)  					& $\mathbf{97.4}$ 	& $\underline{0.030}$			& $\underline{0.277}$ 				& $0.011$			& $\underline{0.206}$ 
			& $\underline{96.2}$ 			& $\underline{0.073}$			& $\underline{0.651}$				& $\underline{0.023}$			& $\underline{0.543}$ \\
			\bottomrule
		\end{tabular}
	}
	\caption{Quantitative comparison of best methods on Apollo 3D Synthetic \cite{genlanenet}. \textbf{Best performance} and \underline{second best} are highlighted.}
	\label{tab:comparison-quant-apollo}
\end{table*}

\section{Experiments}
\label{sec:experiments}
We first describe our experimental setup and then analyze our approach on two 3D lane datasets. 

\subsection{Experimental setup}
\label{subsec:expsetup}
We evaluate our method on two different datasets: OpenLane and Apollo 3D Synthetic - both containing 3D lane ground truth as well as camera parameters per frame.
 
\textbf{OpenLane} \cite{chen2022persformer} is a real-world dataset containing 150,000 images in the training and 40,000 in the test set from 1000 different sequences. In order to evaluate different driving scenarios the test set is divided into different situations, namely \textit{Up \& Down}, \textit{Curve}, \textit{Extreme Weather}, \textit{Night}, \textit{Intersection} and \textit{Merge \& Split}. For ablation studies we use the smaller version OpenLane300 including 300 sequences.

\textbf{Apollo 3D Synthetic} \cite{genlanenet} is a small synthetic dataset, consisting of only 10,500 examples from rather simple scenarios of highway, urban and rural environments. The data is split into three subsets, \textit{(1) Standard} (simple) scenarios, \textit{(2) Rare Scenes} and \textit{(3) Visual Variations}. 

\textbf{Evaluation metrics.} For the quantitative evaluation both datasets utilize the evaluation scheme proposed in \cite{genlanenet}. 

It evaluates the euclidean distance at uniformly distributed points in the range of $0$-$100\,$m along the $y$-direction. 
Based on the mean distance and range, \textit{F1-Score} is computed, as well as the mean $x$- and $z$\textit{-errors} in \textit{near-} ($0$-$40\,$m) and \textit{far-range} ($40$-$100\,$m) to evaluate geometric accuracy. 

\textbf{Baseline.} Our approach builds up on 3D-SpLineNet. Since it was applied on synthetic data only, it showed poor performance on real data. We applied some straight-forward design adaptations - e.g. larger backbone, multi-scale features (see supplementary) - and use this modified 3D-SpLineNet as our baseline (first row \tabref{tab:benefits}).

\textbf{Implementation details.} 
We use input size $360 \times 480$ and adopt the same backbone as in \cite{chen2022persformer} based on a modified EfficientNet \cite{tan2019efficientnet}. We extract four feature maps of resolutions $[\frac{1}{2}, \, \frac{1}{4}, \, \frac{1}{8}, \, \frac{1}{16}]$. 
The final 3D feature map has size $26 \times 16$ with $64$ channels. We use $M=64$ initial line proposals and B-Splines of degree $d=3$ and $K=10$ control points. 
We apply Adam optimizer \cite{kingma2014adam} with an initial learning rate of $2 \times 10^{-4}$ for OpenLane and $10^{-4}$ for Apollo and a dataset specific step-wise scheduler. We train for $30$ epochs on OpenLane and $300$ epochs on Apollo with batch size $16$. 
For more details we refer to the supplementary.

\subsection{Ablation studies}
\tabref{tab:priors} indicates the effect of our proposed prior-based regularization. It is evident that each prior improves the F1-Score as well as geometric errors. While the surface and curvature priors result in better far-range estimates, line parallelism supports X-regression in the near-range. Besides, using surface smoothness loss results in lowest Z-far errors. 
Finally, a combination of priors yields a good balance of F1-Score and geometric errors. 
The positive effect of parallelism is confirmed by \figref{fig:parallelism}, where reinforcing parallel lane structure leads to better estimates in the near-range (a) and far-range (b) compared to the unregularized model. Learning parallel lines also is evidently beneficial in cases of poor visibility (b) and occlusions (a). In the latter case, the regularized model even shows better predictions than the noisy ground truth. This emphasizes the high relevance of priors for more robust behavior for real-world datasets, where 3D ground truth often comes with inaccuracies. 

For the spatial transformation (see \tabref{tab:spatialtrafo}), too low numbers of surface hypotheses result in worse score, presumably as 3D geometry is not captured sufficiently, whereas larger numbers tend to decreasing performance due to the higher complexity. The best F1-Score is obtained with 5 hypotheses, which is chosen for further experiments. While the improvement over IPM is already considerable, we think that with the simplifications of plane hypotheses prevent the component from developing its full potential. We see ways to enhance the 3D transformation even further using more sophisticated spatial representations in future.

The impact of our different contributions is summarized in \tabref{tab:benefits}, where the first row shows our baseline (see Sec.~\ref{subsec:expsetup}). 
More than two percent in F1-Score are gained with our novel lane representation compared to the simplified one from \cite{pittner20233d}. 
Moreover, it is clear that both, the regularization using combined priors and the spatial transformation using 5 hypotheses result in significant improvement. 
Eventually, combining all components yields the best model configuration, which we choose for further evaluation.

\subsection{Evaluation on OpenLane}
On the real-world OpenLane benchmark our model evidently outperforms all other methods with respect to F1-Score as well as geometric errors as shown in \tabref{tab:comparison-quant-ol}.  
Compared to BEV-Lanedet, which achieves a high detection score, our model gains $+1.9\,\%$, while reaching significantly lower geometric errors. In comparison to Anchor3DLane the improvements with respect to X-errors are less substantial, however, our approach surpasses the F1-Score by a large gap of $+6.6\,\%$. 
Analyzing the detection scores among different scenarios, outstanding performance gain is observed on the up- and down-hill test set ($+5.9\,\%$) that highlights the capability of our approach to capture 3D space proficiently, which is supported by the low Z-errors. 

Apart from quantitative results, we show qualitative examples in \figref{fig:sota-comp}. In up-hill scenarios like \figref{fig:sota-comp-b} our model manages to estimate both lateral and height profile accurately, since our assumptions about road surface and line parallelism are satisfied. 
In contrast, PersFormer lacks spatial features and does not use any kind of physical regularization. Consequently, it fails to estimate the 3D lane geometry and even collapses in \figref{fig:sota-comp-c}, whereas our surface and curvature priors always prevent such a behavior. 
Noteworthy is also the top performance on the merges and splits set. This proves that our soft regularization is even capable to handle situations containing non-parallel lines, which is also confirmed by \figref{fig:sota-comp-d}. However, we rarely observe limitations with our formulation for line pairs with a similar orientation but weakly converging course as shown in \figref{fig:sota-comp-e}. In such cases the indicator function might erroneously decide for parallelism loss during training. One possible solution for future work would be to consider ground truth for the indicator function to identify such situations. 

\subsection{Evaluation on Apollo 3D Synthetic}
The Apollo 3D Synthetic dataset is very limited in size and only consists of simple situations in contrast to OpenLane. While we find the results on OpenLane more meaningful, we would like to still provide and discuss the quantitative results on the Apollo dataset. Due to the simplicity of the dataset, our model cannot benefit that significantly from our priors but still achieves competitive results to state of the art with the highest F1-Score on the balanced scenes dataset and comparable error metrics (second best for most errors).
\section{Conclusions and future work}
\label{sec:conclusion}
In this work, we present LaneCPP, a novel approach for 3D lane detection that leverages physical prior knowledge about lane structure and road geometry. 
Our new continuous lane representation overcomes previous deficiencies by allowing arbitrary lane structures and enables us to regularize lane geometry based on analytically formulated priors. 
We further introduce a novel spatial transformation module that models 3D features carefully considering knowledge about road surface geometry. 
In our experiments, we demonstrate state-of-the-art performance on real and synthetic benchmarks. The full capability of our approach is revealed on real-world OpenLane, for which we prove the relevance of priors quantitatively and qualitatively. In future, priors could be individualized for different driving scenarios and might support to learn inter-lane relations to achieve better scene understanding in a global context. We also see ways to leverage the full potential of the spatial transformation by using more sophisticated surface representations.
{
    \small
    \bibliographystyle{ieeenat_fullname}
    \bibliography{survey}
}
\appendix
\setcounter{page}{1}
\maketitlesupplementaryarxiv
\section{Architecture Details}
\label{sec:arch}
In the following section, we provide additional details regarding the model architecture.
\subsection{Backbone}
Similar to \cite{chen2022persformer}, we use a modified version of EfficientNet \cite{tan2019efficientnet} as our backbone. More precisely, we extract a specific layer as the following module's input. Then, several convolution layers are applied, such that the backbone module outputs four different scaled front-view feature maps. Their resolutions are $180 \times 240$, $90 \times 120$, $45 \times 60$, $22 \times 30$. Each of the front-view feature maps is then fed into the spatial transformation module. The total number of parameters of the backbone is $10.28\,$M.

\subsection{Spatial transformation}
The depth branch consists of two convolution layers each with 128 kernels and zero-padding, followed by batch norm and ReLU activation. An additional convolution layer uses $S$ (number of surface hypotheses) kernels of size $1 \times 1$ followed by a channel-wise softmax to obtain the depth distribution. Since the depth distribution should be similar for all front-view feature maps of different scales, only one feature map needs to be propagated through the depth-branch. We use the feature map with lowest resolution $22 \times 30$ and repeat the resulting depth distribution of shape $22 \times 30 \times S$ (with $S$ the number of surface hypotheses) at the neighboring feature cells to match the higher resolutions. Consequently, we obtain depth distributions for all scales of front-view feature maps sharing the same depth information. 

\begin{figure}
	\centering
	\includegraphics[width=.99\linewidth]{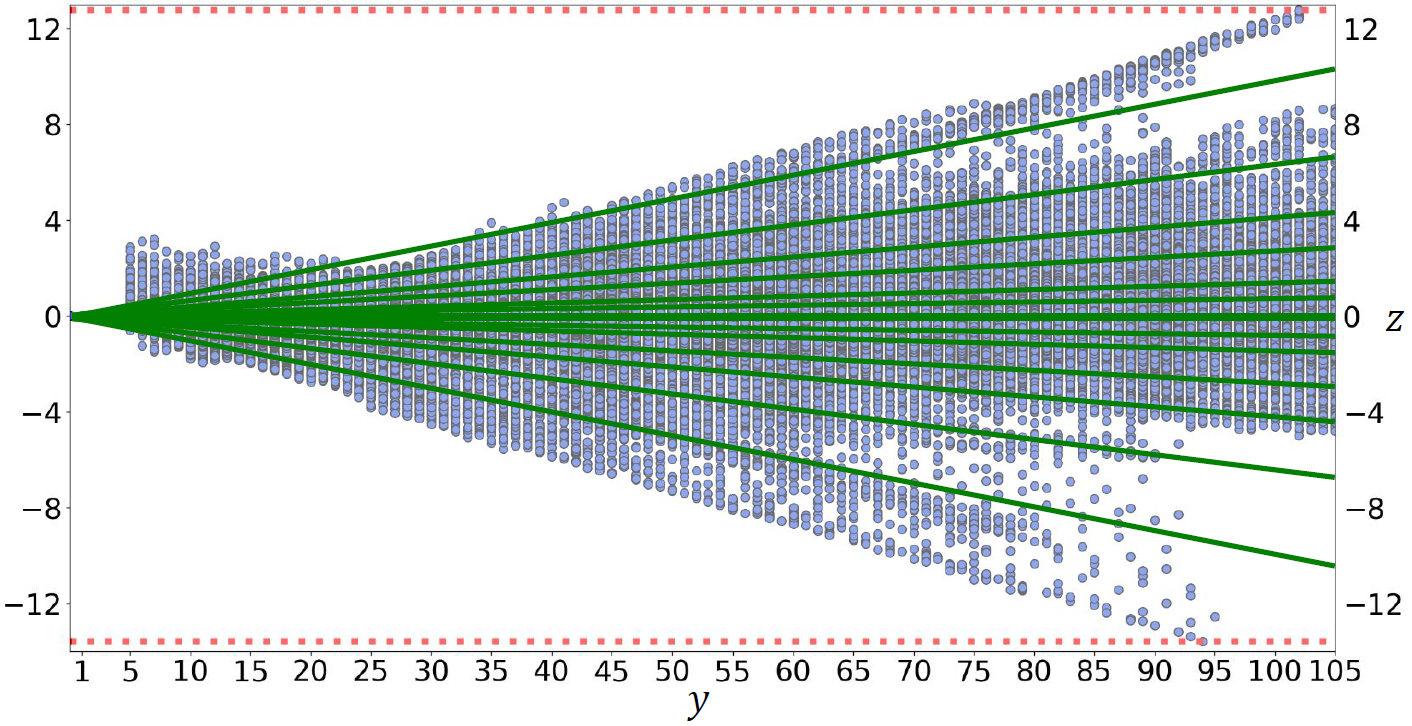}
	\caption{Height distribution ($z$) along the longitudinal direction ($y$) of ground truth line points (blue points) on OpenLane dataset. Height deviations in the near-range (left side) tend to be smaller than in the far-range (right side) spanning a triangle-like region of interest in the $y$-$z$-profile. For the spatial transformation, we sample surface hypotheses (green) of different pitch angles to cover this region.}
	\label{fig:zdist}
\end{figure}
\begin{table}
	\centering
	\resizebox{\columnwidth}{!}{
		\begin{tabular}{|l|l|}
			\hline
			\bf{\# Surface Hypotheses} & \bf{Pitch Angles} \\ 
			\hhline{|=|=|}
			1 			& $\{ 0^\circ \}$ \\
			\hline
			3 			& $\{-2^\circ, 0^\circ, 2^\circ \}$ \\
			\hline
			5 			& $\{-2^\circ, -1^\circ, 0^\circ, 1^\circ, 2^\circ\}$ \\
			\hline
			\multirow{2}{*}{15} 			
						& $\{-5^\circ,  -2^\circ,  -1.7^\circ,  -1.3^\circ,  -1^\circ,  -0.7^\circ,  -0.3^\circ,  0^\circ, $ \\ 
						& $0.3^\circ, 0.7^\circ, 1^\circ, 1.3^\circ, 1.7^\circ, 2^\circ, 5^\circ \}$ \\
			\hline			
			\multirow{4}{*}{27}			
						& $\{-10^\circ, -8.5^\circ, -7^\circ, -5.8^\circ, -4.5^\circ, -3.3^\circ, -2^\circ, $ \\ 
						& $-1.7^\circ, -1.4^\circ, -1^\circ, -0.8^\circ, -0.6^\circ, -0.3^\circ, 0^\circ, $ \\ 
						& $0.3^\circ, 0.6^\circ, 0.8^\circ, 1^\circ, 1.4^\circ, 1.7^\circ, 2^\circ, $ \\ 
						& $3.3^\circ, 4.5^\circ, 5.8^\circ, 7^\circ,  8.5^\circ, 10^\circ \}$ \\
			\hline
	\end{tabular}}
	\caption{Different orientations of surface hypotheses.} \label{tab:surface}
\end{table}

To model the road surface's region of interest, we select surface hypotheses such that the distribution of lane height is covered (see \figref{fig:zdist}). The surface hypotheses are planes crossing the origin of the 3D coordinate system with different orientations with respect to the pitch angle. The different configurations that we use in the experimental section are listed in \tabref{tab:surface}.

After the front-view features are lifted to 3D space they are accumulated on BEV grids. Analogously to the multi-scale front-view feature maps, we also model multi-scale BEV feature maps. The different resolutions are $208 \times 128$, $104 \times 64$, $52 \times 32$, $26 \times 16$.

\subsection{BEV feature fusion}
The BEV feature fusion module consists of convolution layers operating on each scale to down-sample the higher resolutions to the lowest resolution feature map of shape $26 \times 16$. Afterwards, all feature maps are simply concatenated and fed through several layers preserving the resolution. Each contains a convolution with zero-padding, batch norm and ReLU activation. The last convolution layer uses $64$ channels, thus, the input to the detection head is of shape $26 \times 16 \times 64$.

\subsection{Detection head}
\begin{figure}
	\centering
	\includegraphics[width=.99\linewidth]{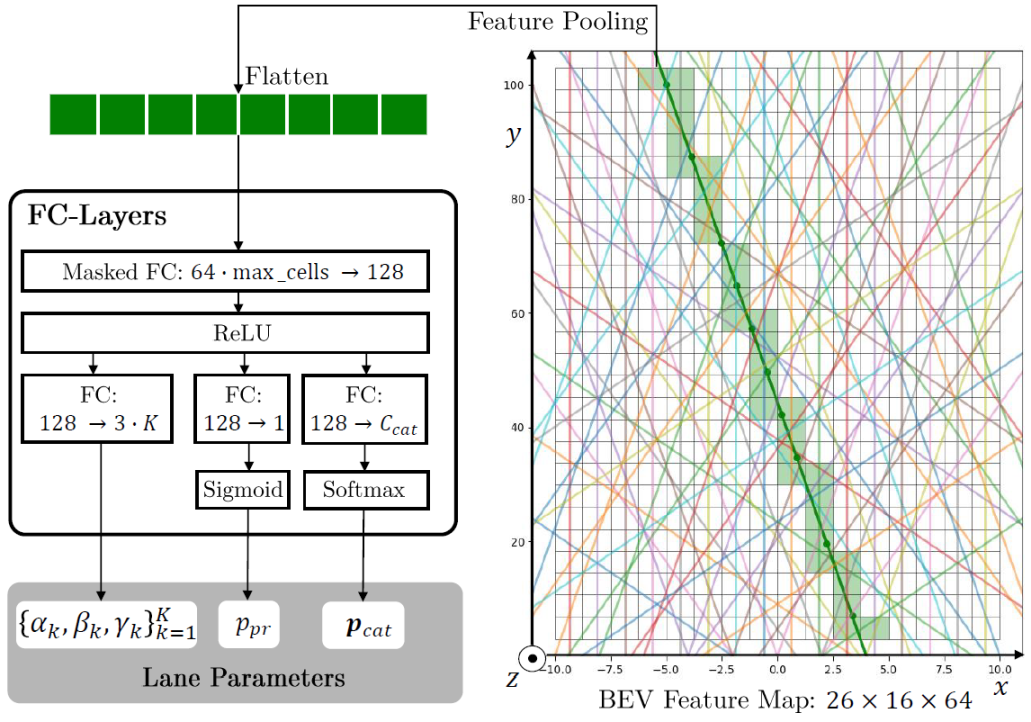}
	\caption{The detection head of our model: First, features are pooled from the BEV feature map for each proposal. Afterwards, pooled features are flattened and fed through several fully-connected (FC) layers, which share weights for all proposals, to finally obtain the lane parameters.}
	\label{fig:head}
\end{figure}

The detection head operates on a BEV feature map of shape $26 \times 16 \times 64$ covering a range of $[-10\,\text{m}, \, 10\,\text{m}]$ in lateral $x$-direction and $[3\,\text{m}, \, 103\,\text{m}]$ in longitudinal $y$-direction. Based on the location of initial line proposals, features are pooled from the BEV feature map for each line proposal as illustrated in \figref{fig:head}. More precisely, we step through a proposal inside the BEV feature grid with a small step size and determine the nearest cells, where the maximum number of cells is limited to $\text{max\_cells}$. We then take the 64-dimensional features of the set of selected cells and flatten it to a feature vector of size $64 \cdot \text{max\_cells}$. If less than $\text{max\_cells}$ are pooled for the proposal, the remaining entries of the feature vector are simply masked out. The resulting feature vector for each line proposal is then propagated through the fully-connected layers as depicted in \figref{fig:head}. Important to notice is also that the fully connected layers share weights among all proposals to learn the same patterns for different line orientations from the BEV feature map. 
Finally, for each proposal the model yields parameters to describe lane line geometry and visibility ($\{\alpha_k, \beta_k, \gamma_k\}_{k=1}^K$), as well as a line presence probability $p_{pr}$ and a probability distribution $\boldsymbol{p}_{cat}$ for different line categories.
\section{Training}
\label{sec:training}
In this section, we describe the details regarding the training procedure. 
\subsection{Initial proposals and Matching}
\begin{figure}
	\centering
	\includegraphics[width=.68\linewidth]{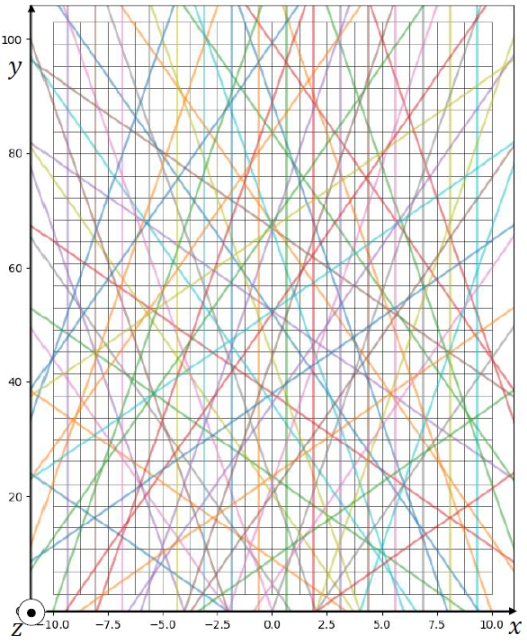}
	\caption{Visualization of different initial line proposals. Colorful lines represent the line proposals. The black lines show the grid of the final BEV feature map.}
	\label{fig:props}
\end{figure}

We use several initial line proposals to cover a wide variety of lane geometries. More precisely, the proposals are straight lines with different orientations and different positions in the $x$-$y$-plane. After investigations of different set configurations, we found the best set of proposals to be the one with $M=64$ proposals that is illustrated in \figref{fig:props}.

The matching of ground truth lines to the line proposals is inspired by \cite{pittner20233d}, which choose the unilateral chamfer distance ($UCD$) as a matching criterion. However, we found that a combination of the unilateral chamfer distance (normalized, thus $UCD \in [0, 1]$) and an orientation cost based on the cosine distance ($CosD \in [0, 1]$) better reflects how well a line proposal $\bar{\boldsymbol{f}}$ resembles a ground truth line described by the set of ground truth points $\mathcal{P}_{GT}$. Thus, the pair-wise matching cost between a proposal with index $i$ (with $i \leq M$) and a ground truth line with index $j$ (with $j \leq M_{GT}$ and $M_{GT}$ the number of ground truth lines) is given as
\begin{align}
	L^{(ij)} =& \lambda_{UCD} \cdot UCD(\bar{\boldsymbol{f}}^{(i)}, \, \mathcal{P}^{(j)}_{GT}) + \\ 
	& \lambda_{CosD} \cdot CosD(\bar{\boldsymbol{f}}^{(i)}, \, \mathcal{P}^{(j)}_{GT}) \, ,
\end{align}
with weights for each cost component $\lambda_{UCD}$ and $\lambda_{CosD}$. 
Computing the cost between each line proposal and each ground truth line then yields a cost matrix of shape \mbox{$M \times M_{GT}$}. Finally, for each ground truth line we assign the proposals with pair-wise cost under a specified matching threshold $L^{(ij)} < L_{thr}$.

\subsection{Losses and ground truth}
We provide more details regarding losses and ground truth.

\noindent \textbf{Indicator function for prior regularization.} The parallelism loss uses an indicator function $\mathbbm{1}_{\boldsymbol{p}}^{(ij)}$ deciding, whether the loss is applied to the point pair consisting of point $\boldsymbol{p}$ on line $i$ and the best matching point in normal direction $\boldsymbol{p}^*$ on line $j$. The indicator function is defined as
\begin{align}
	\mathbbm{1}_{\boldsymbol{p}}^{(ij)} = \begin{cases}
	1 \,  & \text{if} \qquad OD_{\boldsymbol{p}^*}^{(ij)} < OD_{thr} \;\; \text{and} \;\; \sigma^{(ij)} < \sigma_{thr} \, , \\
	0 \,  & \text{else} \, .
	\end{cases} \label{eq:indicator}
\end{align}
As \cref{eq:indicator} shows, the parallelism criterion holds if two conditions are fulfilled. 
The first condition $OD_{\boldsymbol{p}^*}^{(ij)} < OD_{thr}$ takes into account the orthogonal distance ($OD$) of the best matching point $\boldsymbol{p}^*$ on line $j$ to the normal plane spanned by the tangent $\mathbf{T}^{(i)}(t_{\boldsymbol{p}})$ at point $\boldsymbol{p}$ on line $i$, which is given as 
\begin{align}
	OD_{\boldsymbol{p}^*}^{(ij)} = \mathbf{T}^{(i)}(t_{\boldsymbol{p}})^T \cdot \big(\boldsymbol{f}^{(j)}({t_{{\boldsymbol{p}}^*}}) - \boldsymbol{f}^{(i)}(t_{\boldsymbol{p}})\big) \, .
\end{align}
Hence, only point pairs are considered for the parallelism loss, which actually lie in opposite normal direction. This is implied by the orthogonal distance having a small enough value, i.e. if the value is lower than a certain threshold $OD_{thr}$. For instance, if two neighboring lines have different ranges, the non-overlapping range has no neighbor points that have an orthogonal distance smaller than the threshold. Thus, the condition ensures that only point pairs are considered, which are actual neighbors in normal direction.

The second condition $\sigma^{(ij)} < \sigma_{thr}$ guarantees that parallelism is not reinforced for line pairs, which presumably belong to lanes of different orientations, e.g. for merge and split scenarios. The distinction between parallel and non-parallel line pairs can be determined by evaluating the standard deviation $\sigma^{(ij)}$ of the euclidean distances $D_{\boldsymbol{p}}^{(ij)}$ of point pairs of neighboring lines $i$ and $j$. The standard deviation is defined as
\begin{align}
	\sigma^{(ij)} =& \sqrt{\frac{1}{| \mathcal{P}^{(i)} |} \sum_{\boldsymbol{p} \in \mathcal{P}^{(i)}} D_{\boldsymbol{p}}^{(ij)} - \bar{D}^{(ij)}} \, , \, \mathrm{where} \\ \bar{D}^{(ij)} =& \frac{1}{| \mathcal{P}^{(i)} |} \sum_{\boldsymbol{p} \in \mathcal{P}^{(i)}} D_{\boldsymbol{p}}^{(ij)} \, ,
\end{align}
and the euclidean distance for one point pair as $D_{\boldsymbol{p}}^{(ij)} = \big|\big| {\boldsymbol{f}^{(i)}(t_{\boldsymbol{p}}) - \boldsymbol{f}^{(j)}(t_{\boldsymbol{p}^*})} \big|\big|_2$. For lines of different orientations (as for merging and splitting lines) this standard deviation is rather high and more likely surpasses the threshold $\sigma_{thr}$ in contrast to lines belonging to the same lane, where $\sigma^{(ij)}$ is rather small. 

\begin{figure}
	\centering
	\begin{subfigure}{.43\textwidth}
		\centering
		\includegraphics[width=.99\linewidth]{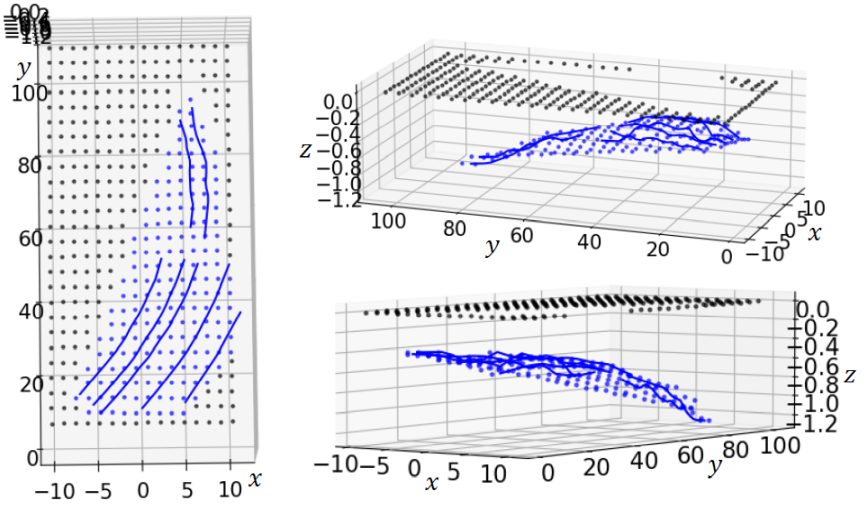}
		\caption{Down-hill scenario}
		\label{fig:gridsurface-down}
	\end{subfigure}

	\begin{subfigure}{.43\textwidth}
		\centering
		\includegraphics[width=.99\linewidth]{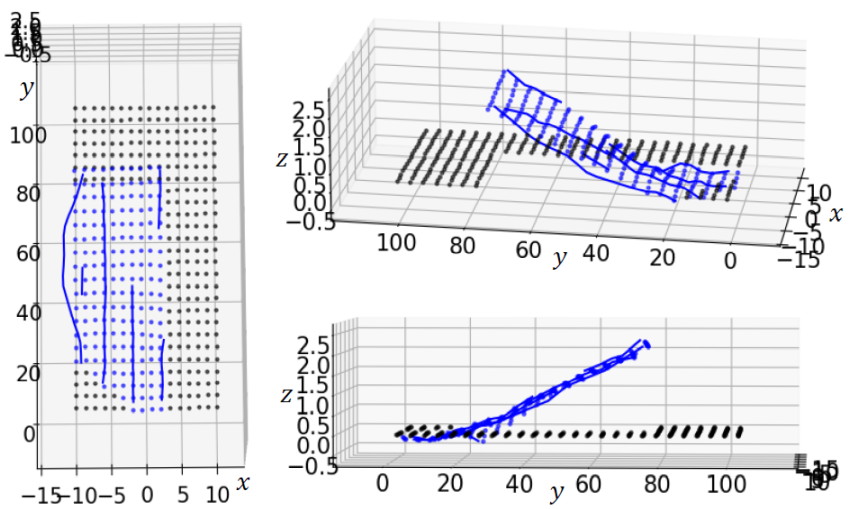}
		\caption{Up-hill scenario}
		\label{fig:gridsurface-up}
	\end{subfigure}
	\caption{Examples of the surface ground truth generation. Ground truth lines are visualized as blue lines and height ground truth per cell as blue dots. The black dots correspond to cells outside the convex hull of 3D lines and are not considered for the surface loss.}
	\label{fig:surfacegt}
\end{figure}

\noindent \textbf{Ground truth generation for surface loss.} For the surface loss computation, height ground truth $\hat{z}_{uv}$ needs to be provided on the $X \times Y$ BEV grid. We approximate this surface ground truth by interpolation of the 3D lane ground truth. For this, we simply compute the convex hull of ground truth lines and interpolate the height value at each cell inside the convex hull. Only cells inside the convex hull are considered for the surface loss, whereas cells outside the convex hull are simply masked out. This is reflected by the indicator function $\mathbbm{1}_{uv}$, hence $\mathbbm{1}_{uv} = 1$ if cell $(u,v)$ is inside the hull, else $\mathbbm{1}_{uv} = 0$. The result of the grid-wise height ground truth generation is visualized in \figref{fig:surfacegt} for an up-hill and a down-hill scenario.

\noindent \textbf{Lane presence and category classification losses.} For both classification losses, we apply focal loss \cite{lin2017focal}. For line presence, which only considers the two classes present and not present, the loss is given as
\begin{align}
\mathcal{L}_{pr} =& - \frac{1}{M} \sum_{i=1}^{M} \Big( \hat{p}_{pr}^{(i)} \cdot \big(1-p_{pr}^{(i)}\big)^{\gamma_{f}} \cdot \log\big(p_{pr}^{(i)}\big) + \\ 
&\big(1 - \hat{p}_{pr}^{(i)}\big) \cdot \big(p_{pr}^{(i)}\big)^{\gamma_{f}} \cdot \log\big(1 - p_{pr}^{(i)}\big) \Big) \, , \label{eq:linepresence}
\end{align}
with predicted line presence probability $p_{pr}^{(i)}$ for line $i$ and line presence ground truth $\hat{p}_{pr}^{(i)} = \{0, 1\}$. $\gamma_{f} \geq 0$ denotes the focusing parameter introduced in \cite{lin2017focal} to handle class imbalance.

The category classification loss is applied for datasets, which provide lane category information in the ground truth. Analogously to \cref{eq:linepresence}, the loss is given as
\begin{align}
\mathcal{L}_{cat} =& - \frac{1}{M} \sum_{i=1}^{M} \frac{1}{C_{cat}} \sum_{c=1}^{C_{cat}} \Big( \hat{\boldsymbol{p}}_{cat}^{(i)}[c] \cdot \\ & \big(1-\boldsymbol{p}_{cat}^{(i)}[c]\big)^{\gamma_{f}} \cdot \log\big(\boldsymbol{p}_{cat}^{(i)}[c]\big) \Big) \, ,
\end{align}
with the predicted category probability vector $\boldsymbol{p}_{cat}^{(i)} \in \mathbb{R}^{C_{cat}}$, which represents the categorical distribution for line $i$, and the ground truth one-hot vector $\hat{\boldsymbol{p}}_{cat}^{(i)} \in \{0, 1\}^{C_{cat}}$. Moreover, $\boldsymbol{p}_{cat}^{(i)}[c]$ denotes the $c^{\text{th}}$ entry of the vector $\boldsymbol{p}_{cat}^{(i)}$.

\begin{figure}
	\centering
	\begin{subfigure}{.24\textwidth}
		\centering
		\includegraphics[width=.99\linewidth]{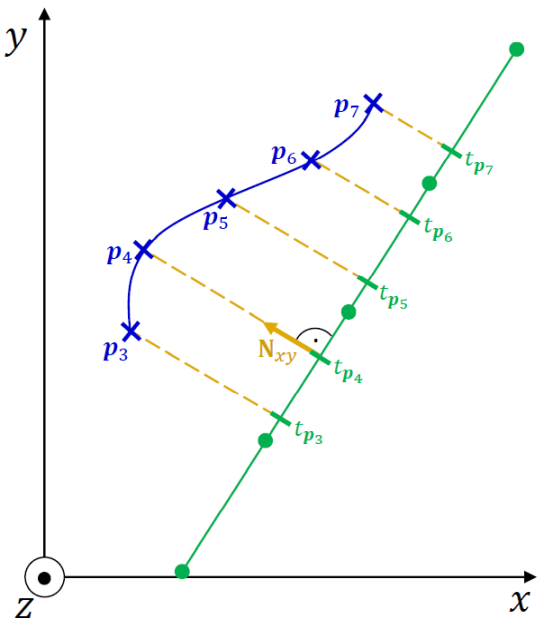}
		\caption{Regression}
		\label{fig:tcompreg}
	\end{subfigure}%
	\begin{subfigure}{.24\textwidth}
		\centering
		\includegraphics[width=.99\linewidth]{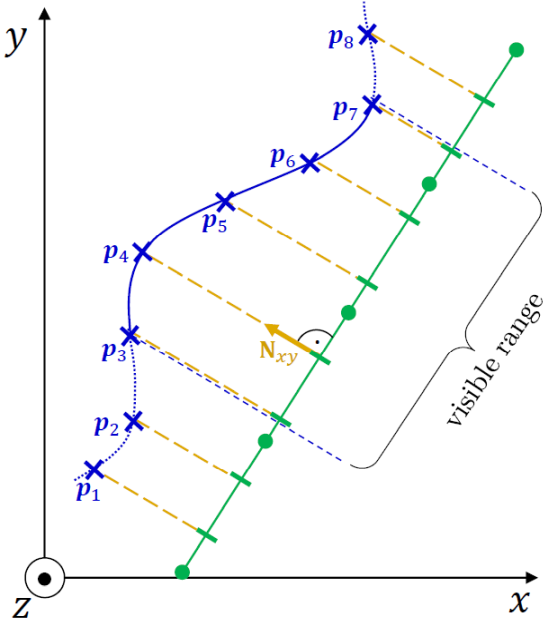}
		\caption{Visibility}
		\label{fig:tcompvis}
	\end{subfigure}
	\caption{Projection of \textcolor{blue}{ground truth} points $\boldsymbol{p}$ onto \textcolor{darkgreen}{line proposal} in \textcolor{orange}{normal direction} to obtain curve arguments $t_{\boldsymbol{p}}$. For regression (a) only visible points are considered (continuous lines), for visibility (b) all points are taken into account, where invisible points are marked with dashed lines.}
	\label{fig:tcomp}
\end{figure}

\noindent \textbf{Regression loss.} For both, the regression and visibility loss, the curve argument $t_{\boldsymbol{p}}$ has to be determined for a respective point in the ground truth $\boldsymbol{p} \in \mathcal{P}_{GT}$. Since our model learns to predict orthogonal offsets from the assigned line proposal, the points are projected orthogonal onto the line proposal as illustrated in \figref{fig:tcomp}. After having obtained the curve arguments in orthogonal direction, the regression loss for a line proposal $i$ is given as
\begin{align}
\mathcal{L}_{reg}^{(i)} =& \frac{1}{|\mathcal{P}_{GT}^{(i)}|} \sum_{\boldsymbol{p} \in \mathcal{P}_{GT}^{(i)}} \hat{v}_{\boldsymbol{p}}^{(i)} \cdot \Big|\Big|
\boldsymbol{w} \odot \Big( \boldsymbol{f}^{(i)}(t_{\boldsymbol{p}})
-
\begin{pmatrix}
\hat{x}_{\boldsymbol{p}}^{(i)}   \\
\hat{y}_{\boldsymbol{p}}^{(i)}	\\
\hat{z}_{\boldsymbol{p}}^{(i)}
\end{pmatrix}
\Big) \Big|\Big|_1 \label{eq:regloss}
\end{align}
with $\hat{v}_{\boldsymbol{p}}^{(i)}$ the ground truth visibility information and $(\hat{x}_{\boldsymbol{p}}^{(i)}, \hat{y}_{\boldsymbol{p}}^{(i)}, \hat{z}_{\boldsymbol{p}}^{(i)})^T$ the 3D position of a ground truth point $\boldsymbol{p}$ on line $i$. $\boldsymbol{w} \in \mathbb{R}^3$ is a vector with weighting factors for each 3D component providing for a more balanced regression in each dimension. As shown in \cref{eq:regloss} and illustrated in \figref{fig:tcompreg}, only visible points are utilized. 
The total regression loss for all lines is given as
\begin{align}
	\mathcal{L}_{reg} = \frac{1}{M} \sum_{i=1}^{M} \, \hat{p}_{pr}^{(i)} \cdot \mathcal{L}_{reg}^{(i)} \, .
\end{align}
For completeness, we also provide the visibility loss for each line as
\begin{align}
	\mathcal{L}_{vis}^{(i)} =& - \frac{1}{|\mathcal{P}_{GT}^{(i)}|} \sum_{\boldsymbol{p} \in \mathcal{P}_{GT}^{(i)}} \hat{v}_{\boldsymbol{p}}^{(i)} \cdot \log\big(\sigma \big(v^{(i)}(t_{\boldsymbol{p}})\big)\big) + \\ 
&(1- \hat{v}_{\boldsymbol{p}}^{(i)}) \cdot \log\big(1 - \sigma \big(v^{(i)}(t_{\boldsymbol{p}})\big)\big) \, .
\end{align}
As illustrated in \figref{fig:tcompvis}, all points from the ground truth line are considered. The total visibility loss is then given as
\begin{align}
\mathcal{L}_{vis} = \frac{1}{M} \sum_{i=1}^{M} \, \hat{p}_{pr}^{(i)} \cdot \mathcal{L}_{vis}^{(i)} \, .
\end{align}

\section{Additional implementation details}
In the following, we provide more implementation details.
\subsection{Matching} 
The weights for the matching cost are $\lambda_{UCD} = 0.5$ and $\lambda_{CosD} = 0.5$, and the distance threshold $L_{thr} = 0.4$.
\subsection{Losses}
The weights for the different losses are $\lambda_{pr} = 20$, $\lambda_{cat} = 2$, $\lambda_{reg} = 0.5$, $\lambda_{par}=10$, $\lambda_{sm}=0.01$, $\lambda_{curv}=1$, $\lambda_{prior}=1$, $\lambda_{surf}=0.1$. The focusing parameter for the classification losses is $\gamma_f = 6.0$ and the vector to weight each dimension for the regression loss is $\boldsymbol{w} = (2, 10, 1)^T$. The thresholds for the indicator function used for the prior losses are $\sigma_{thr}=2\,\text{m}$ and $OD_{thr}=1\,\text{m}$ and the thresholds for the maximum curvatures are $\kappa_{xy}=5$ and $\kappa_{z}=0.1$. The set of ground truth points considered for the visibility and regression losses has size $|\mathcal{P}_{GT}|=20$. For the parallelism and surface smoothness loss we sample $|\mathcal{P}|=20$ points from the predictions and $|\mathcal{P}|=100$ points for the curvature loss.
\subsection{Training procedure}
In the training, we use Adam optimizer \cite{kingma2014adam}, with an initial learning rate of $2 \cdot 10^{-4}$ for OpenLane and $10^{-4}$ for Apollo. We use a dataset specific scheduler: We train for $30$ epochs on OpenLane, where the learning rate is decreased to $5 \cdot 10^{-5}$ after $27$ epochs, and for $300$ epochs on Apollo, where the learning rate is divided by two every 100 epochs. 
\subsection{Others}
The maximum number of cells used for feature pooling in the detection head is $\text{max\_cells} = 64$.
\section{Additional results}
\label{sec:moreresults}
In this section, we provide additional quantitative and qualitative results.
\subsection{Ablation studies}
\tabref{tab:add-adaptations} shows the performance of 3D-SpLineNet \cite{pittner20233d} on OpenLane300 and the effect of different design adaptations. It is clearly evident that these modifications result in large improvements that were necessary to make the approach applicable to real-world data. 
\begin{table}
	\centering
	\resizebox{\columnwidth}{!}{
		\begin{tabular}{l||c|c|c|c}
			\toprule
			\bf{Config} & \bf{3D-SpLineNet} & \bf{+BB} & \bf{+BB+MS} & \bf{+BB+MS+FP} \\ 
			\hline
			\bf{F1(\%)}$\uparrow$	& $50.9$ 	& $53.7$ 	& $58.7$ 	& $62.9$ \\
			\bottomrule
	\end{tabular}}
	\caption{Performance on OpenLane300 of 3D-SpLineNet baseline and architecture adaptations, i.e. larger backbone (BB), multi-scale features (MS) and feature pooling in detection head (FP).} \label{tab:add-adaptations}
\end{table}
\begin{table}
	\centering
	\resizebox{\columnwidth}{!}{
		\begin{tabular}{c|c|ccccc}
			\toprule
			\multirow{2}{*}{\bf{Uniform}} & \bf{Sampling Rate} & \bf{1} & \bf{3} & \bf{5} & \bf{15} & \bf{27} \\
			& \bf{F1-Score(\%)}$\uparrow$	& 15.4 & 30.9 & 39.6 & 48.4 & 51.1 \\
			\hhline{=======}
			\multirow{2}{*}{\bf{Surface H.}} & \bf{Sampling Rate} & \bf{1} & \bf{3} & \bf{5} & \bf{15} & \bf{27} \\
			& \bf{F1-Score(\%)}$\uparrow$	&  65.0 & 65.9 & 66.6 & 66.1 & 66.0 \\
			\bottomrule
		\end{tabular}
	}
	\caption{Effect of the sampling strategy used in the spatial transformation on OpenLane300. Uniform ray sampling is compared to samples obtained from intersections of rays with surface hypotheses.} \label{tab:add-spatial}
\end{table}

In \tabref{tab:add-spatial} we compare two different strategies to draw samples from the camera rays to investigate the effect of using priors in form of surface hypotheses for this component. The samples determine the frustum-like pseudo point cloud in 3D space as described in Sec.~3.3 in the main paper. For the uniform sampling (comparable to \cite{philion2020lift}), the samples are drawn along the rays with equal step size in the range $[3\,\textrm{m}, 110\,\textrm{m}]$ to guarantee that the whole space of interest is covered. We compare this method to our sampling based on prior-incorporated surface hypotheses as proposed and described in the main paper. As shown in \tabref{tab:add-spatial}, the performance gaps between the two strategies are significant. This highlights the importance of modeling geometry-aware 3D features by generating samples in the space of interest using knowledge about the surface geometry. The differences in F1-Score for varying sampling rates also imply that a uniform sampling strategy requires high sampling rates to achieve comparable performance. In contrast, using surface hypotheses, lower sampling rates are sufficient which keeps the computational costs lower.

\subsection{Quantitative results}
\label{sec:morequant}
In \tabref{tab:comparison-quant-ol-more} we report the detailed evaluation metrics of our best performing LaneCPP model for the different scenarios on OpenLane. We provide geometric errors, as well as F1-Score, precision, recall and categorical accuracy.

Besides, we provide a more detailed evaluation on the Apollo 3D Synthetic dataset on all three test sets as shown in \tabref{tab:comparison-quant-sim-more}. 

\begin{table*}[tb]
	\centering
		\begin{tabular}{l||ccc|c|cccc}
			\toprule
			\multirow{2}{*}{\bf{Scenario}} & \multirow{2}{*}{\bf{F1(\%)}$\uparrow$} & \multirow{2}{*}{\bf{P(\%)}$\uparrow$} & \multirow{2}{*}{\bf{R(\%)}$\uparrow$} &	\bf{Categorical} & \multicolumn{2}{c}{\bf{X-error (m)}$\downarrow$} & \multicolumn{2}{c}{\bf{Z-error (m)}$\downarrow$} \\ 
			&  & & & \bf{Accuracy(\%)}$\uparrow$ & \bf{near} & \bf{far} & \bf{near} & \bf{far} \\
			\hhline{=||===|=|====}
			\bf{Up \& Down} 				& $53.6$ 		& $58.4$			& $49.5$			& $90.0$ 		& $0.338$ 			& $0.433$ 		& $0.122$		& $0.188$ 	\\ 
			\bf{Curve} 						& $64.4$ 		& $67.7$			& $61.4$			& $91.1$ 		& $0.283$ 			& $0.441$ 		& $0.075$		& $0.117$ 	\\ 
			\bf{Extreme Weather} 			& $56.7$ 		& $63.4$			& $51.2$			& $88.8$ 		& $0.333$ 			& $0.253$ 		& $0.081$		& $0.113$ 	\\ 
			\bf{Night} 						& $54.9$ 		& $60.6$			& $50.2$			& $82.9$ 		& $0.318$ 			& $0.323$ 		& $0.104$		& $0.166$ 	\\
			\bf{Intersection}				& $52.0$		& $56.6$			& $48.1$			& $84.7$		& $0.316$ 			& $0.343$		& $0.099$		& $0.140$	\\
			\bf{Merge \& Split}				& $58.7$		& $63.2$			& $54.8$			& $86.0$		& $0.284$			& $0.330$		& $0.066$		& $0.105$	\\
			\bf{All}						& $60.3$		& $64.7$			& $56.5$			& $87.1$		& $0.264$			& $0.310$		& $0.077$		& $0.117$	\\
			\bottomrule
	\end{tabular}
	\caption{Detailed quantitative evaluation of our LaneCPP for different scenarios on OpenLane \cite{chen2022persformer}.}
	\label{tab:comparison-quant-ol-more}
\end{table*}

\begin{table*}[tb]
	\begin{center}
		\begin{tabular}{llcccccc}
			\toprule
			\multirow{2}{*}{\bf{Scenario}} & \multirow{2}{*}{\bf{Method}} & \multirow{2}{*}{\bf{F1-Score(\%)}$\uparrow$} & \multirow{2}{*}{\bf{AP(\%)}$\uparrow$} & \multicolumn{2}{c}{\bf{X-error (m)}$\downarrow$} & \multicolumn{2}{c}{\bf{Z-error (m)}$\downarrow$} \\
			& & & & \bf{near} & \bf{far} & \bf{near} & \bf{far} \\
			\hhline{========}
			& 
			3D-LaneNet \cite{3dlanenet} 					& $86.4$				& $89.3$  				& $0.068$ 				& $0.477$ 				& $0.015$ 				& $\mathbf{0.202}$ \\
			& Gen-LaneNet \cite{genlanenet} 				& $88.1$ 				& $90.1$ 				& $0.061$ 				& $0.496$ 				& $0.012$ 				& $0.214$ \\
			& 3D-LaneNet (1/att) \cite{jin2021robust}		& $91.0$ 				& $93.2$				& $0.082$ 				& $0.439$ 				& $0.011$ 				& $0.242$ \\
			& Gen-LaneNet (1/att) \cite{jin2021robust}		& $90.3$ 				& $92.4$ 				& $0.08$ 				& $0.473$ 				& $0.011$ 				& $0.247$ \\
			& CLGO \cite{liu2022learning} 					& $91.9$ 				& $94.2$ 				& $0.061$ 				& $0.361$ 				& $0.029$ 				& $0.250$ \\ 
			\it{Balanced} 		
			& GP \cite{li2022reconstruct} 					& $91.9$ 				& $93.8$				& $0.049$ 				& $0.387$ 				& $\mathbf{0.008}$ 		& $0.213$ \\
			\it{Scenes} 
			& PersFormer \cite{chen2022persformer} 			& $92.9$ 				& $-$ 					& $0.054$ 				& $0.356$ 				& $0.010$ 				& $0.234$ \\
			& 3D-SpLineNet \cite{pittner20233d}				& $96.3$				& $\underline{98.1}$	& $0.037$				& $0.324$				& $\underline{0.009}$				& $0.213$ \\
			& CurveFormer \cite{bai2023curveformer}			& $95.8$ 				& $97.3$ 				& $0.078$ 				& $0.326$ 				& $0.018$ 				& $0.219$ \\
			& BEV-LaneDet \cite{wang2023bev}  				& $\underline{96.9}$ 	& $-$ 					& $\mathbf{0.016}$ 		& $\mathbf{0.242}$ 		& $0.02$ 				& $0.216$ \\
			& Anchor3DLane \cite{huang2023anchor3dlane}		& $95.4$				& $97.1$ 				& $0.045$				& $0.300$				& $0.016$ 				& $0.223$ \\
			\rowcolor{Gray} & LaneCPP 		 				& $\mathbf{97.4}$ 		& $\mathbf{99.5}$ 		& $\underline{0.030}$	& $\underline{0.277}$ 	& $0.011$				& $\underline{0.206}$ \\
			\hline
			& 
			3D-LaneNet \cite{3dlanenet} 					& $72.0$				& $74.6$  				& $0.166$ 				& $0.855$ 				& $0.039$ 				& $\mathbf{0.521}$ \\
			& Gen-LaneNet \cite{genlanenet} 				& $78.0$ 				& $79.0$ 				& $0.139$ 				& $0.903$ 				& $0.030$ 				& $0.539$ \\
			& 3D-LaneNet (1/att) \cite{jin2021robust}		& $84.1$ 				& $85.8$				& $0.289$ 				& $0.925$ 				& $0.025$ 				& $0.625$ \\
			& Gen-LaneNet (1/att) \cite{jin2021robust}		& $81.7$				& $83.2$				& $0.283$ 				& $0.915$		 		& $0.028$ 				& $0.653$ \\
			& CLGO \cite{liu2022learning} 					& $86.1$ 				& $88.3$ 				& $0.147$	 			& $0.735$ 				& $0.071$ 				& $0.609$ \\
			\it{Rare}
			& GP \cite{li2022reconstruct} 					& $83.7$ 				& $85.2$				& $0.126$ 				& $0.903$ 				& $0.023$ 				& $0.625$ \\
			\it{Scenes}
			& PersFormer \cite{chen2022persformer} 			& $87.5$ 				& $-$ 					& $0.107$ 				& $0.782$ 				& $0.024$ 				& $0.602$ \\
			& 3D-SpLineNet \cite{pittner20233d}				& $92.9$				& $94.8$				& $0.077$				& $0.699$				& $\mathbf{0.021}$		& $0.562$ \\
			& CurveFormer \cite{bai2023curveformer}			& $95.6$ 				& $\underline{97.1}$	& $0.182$ 				& $0.737$ 				& $0.039$ 				& $0.561$ \\
			& BEV-LaneDet \cite{wang2023bev}  				& $\mathbf{97.6}$ 		& $-$ 					& $\mathbf{0.031}$ 		& $\mathbf{0.594}$ 		& $0.040$ 				& $0.556$ \\
			& Anchor3DLane \cite{huang2023anchor3dlane}		& $94.4$				& $95.9$ 				& $0.082$				& $0.699$				& $0.030$ 				& $0.580$ \\
			\rowcolor{Gray} & LaneCPP 						& $\underline{96.2}$ 	& $\mathbf{98.6}$		& $\underline{0.073}$	& $\underline{0.651}$	& $\underline{0.023}$	& $\underline{0.543}$ \\
			\hline
			& 
			3D-LaneNet \cite{3dlanenet} 					& $72.5$				& $74.9$  				& $0.115$ 				& $0.601$ 				& $0.032$ 				& $0.230$ \\
			& Gen-LaneNet \cite{genlanenet} 				& $85.3$ 				& $87.2$ 				& $0.074$ 				& $0.538$ 				& $0.015$ 				& $0.232$ \\
			& 3D-LaneNet (1/att) \cite{jin2021robust}		& $85.4$ 				& $87.4$				& $0.118$ 				& $0.559$ 				& $0.018$ 				& $0.290$ \\
			& Gen-LaneNet (1/att) \cite{jin2021robust}		& $86.8$ 				& $88.5$ 				& $0.104$ 				& $0.544$ 				& $0.016$ 				& $0.294$ \\
			& CLGO \cite{liu2022learning} 					& $87.3$ 				& $89.2$ 				& $0.084$ 				& $0.464$ 				& $0.045$ 				& $0.312$ \\
			\it{Visual}
			& GP \cite{li2022reconstruct} 					& $89.9$ 				& $92.1$				& $0.060$ 				& $0.446$ 				& $\mathbf{0.011}$ 				& $0.235$ \\
			\it{Variations} 
			& PersFormer \cite{chen2022persformer} 			& $89.6$ 				& $-$ 					& $0.074$ 				& $0.430$ 				& $0.015$ 				& $0.266$ \\
			& 3D-SpLineNet \cite{pittner20233d}				& $91.3$				& $\underline{93.1}$	& $0.069$				& $0.468$				& $\underline{0.013}$				& $0.248$ \\
			& CurveFormer \cite{bai2023curveformer}			& $90.8$ 		 		& $93.0$ 				& $0.125$				& $0.410$ 				& $0.028$	 			& $0.254$ \\
			& BEV-LaneDet \cite{wang2023bev}  				& $\mathbf{95.0}$ 		& $-$ 					& $\mathbf{0.027}$ 		& $\mathbf{0.320}$ 		& $0.031$ 				& $0.256$ \\
			& Anchor3DLane \cite{huang2023anchor3dlane}		& $\underline{91.8}$	& $92.5$ 				& $\underline{0.047}$	& $\underline{0.327}$	& $0.019$ 				& $\mathbf{0.219}$ \\
			\rowcolor{Gray} & LaneCPP 						& $90.4$ 				& $\mathbf{93.7}$ 		& $0.054$				& $\underline{0.327}$	& $0.020$					& $\underline{0.222}$	  \\
			\bottomrule
		\end{tabular}
		\caption{Quantitative evaluation on Apollo 3D Synthetic \cite{genlanenet}. \textbf{Best performance} and \underline{second best} are highlighted.}
		\label{tab:comparison-quant-sim-more}
	\end{center}
\end{table*}

\subsection{Qualitative results}
\label{sec:morequal}
We show additional qualitative results on OpenLane in \figref{fig:supp-qual-ol-1}. Considering the top rows, it is clearly evident in all examples that our LaneCPP detects lanes more accurately compared to 3D-SpLineNet, which performs poorly on real-world data. The bottom row shows a direct comparison of LaneCPP and PersFormer. Particularly in curves (\figref{fig:supp-qual-ol-1a} - \figref{fig:supp-qual-ol-1b}) and up- or down-hill scenarios (\figref{fig:supp-qual-ol-2a} - \figref{fig:supp-qual-ol-2c}) our model shows high-quality detections compared to PersFormer. For the intersection scenario (\figref{fig:supp-qual-ol-1c}) with many different line instances, LaneCPP shows overall good results but still leaves room for improvement with respect to geometrical precision. A possible solution to improve the behavior in such cases could be to model lane line relations explicitly to better capture global context as mentioned in our future work section. Moreover, we prove that our model is able to classify line categories accurately as illustrated in the middle row plots.

\begin{figure*}
	\centering
	\begin{subfigure}{.97\textwidth}
		\centering
		\includegraphics[width=.99\linewidth]{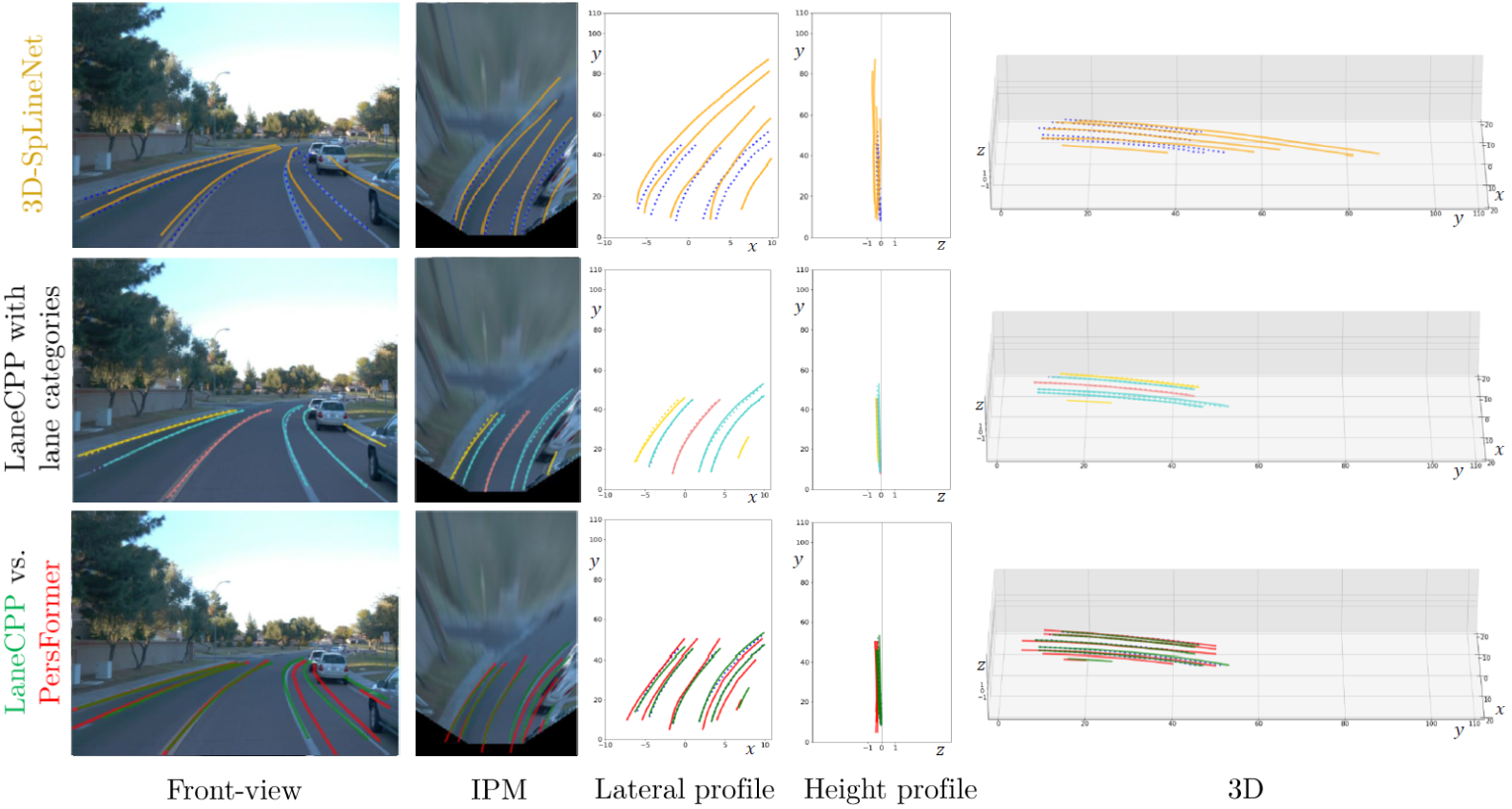}
		\caption{}
		\label{fig:supp-qual-ol-1a}
	\end{subfigure}
	\begin{subfigure}{.97\textwidth}
		\centering
		\includegraphics[width=.99\linewidth]{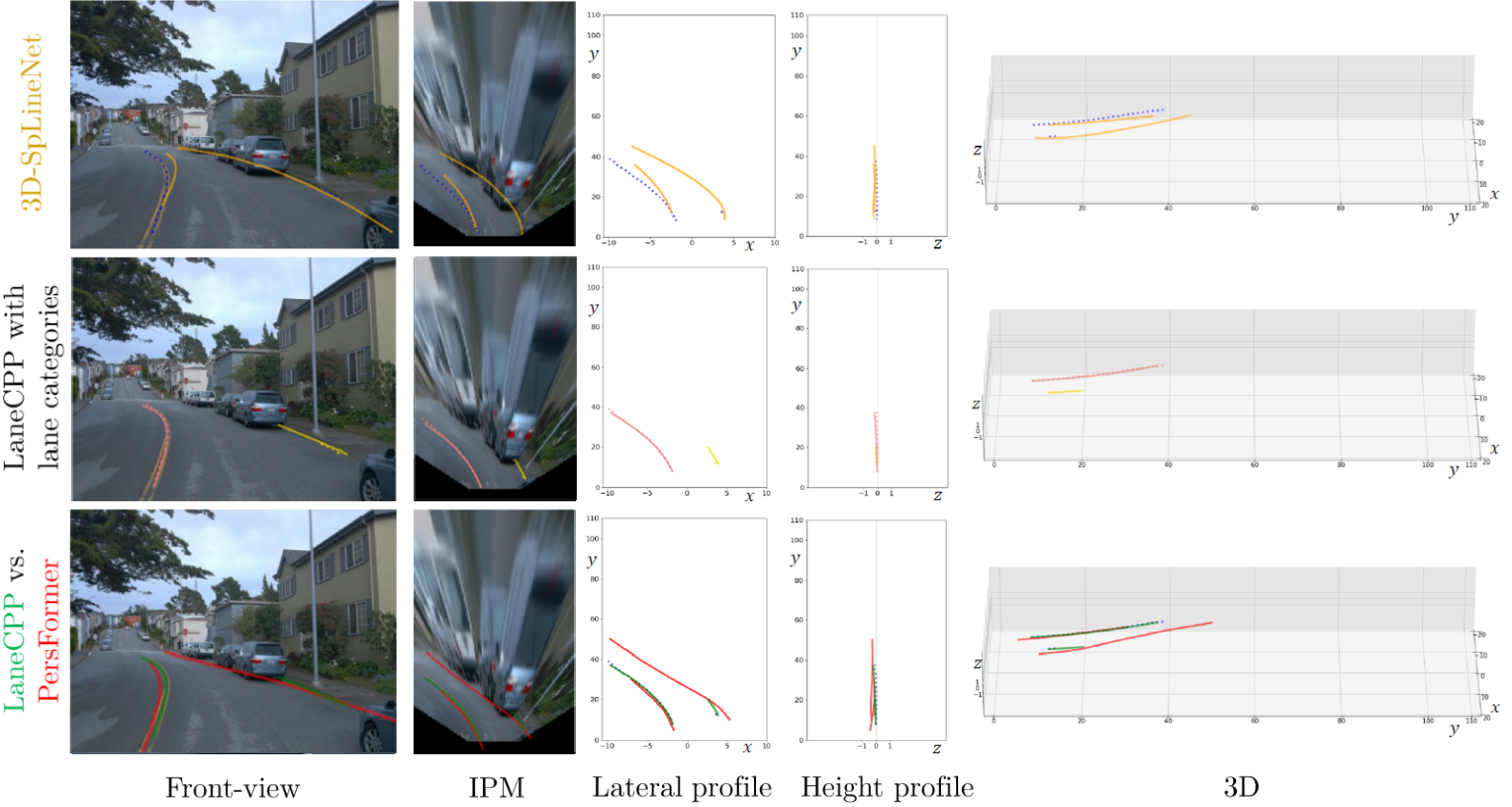}
		\caption{}
		\label{fig:supp-qual-ol-1b}
	\end{subfigure}
	\caption{Additional qualitative evaluation on OpenLane \cite{chen2022persformer} test set (1/3). Top row shows \textcolor{orange}{3D-SpLineNet} baseline compared to \textcolor{blue}{ground truth}. Middle row shows LaneCPP with different lane categories illustrated in different colors and ground truth in dashed lines. Bottom row shows direct comparison of \textcolor{darkgreen}{LaneCPP} and \textcolor{red}{PersFormer*}.}
	\label{fig:supp-qual-ol-1}
\end{figure*}

\begin{figure*}\ContinuedFloat
	\centering
	\begin{subfigure}{.97\textwidth}
		\centering
		\includegraphics[width=.99\linewidth]{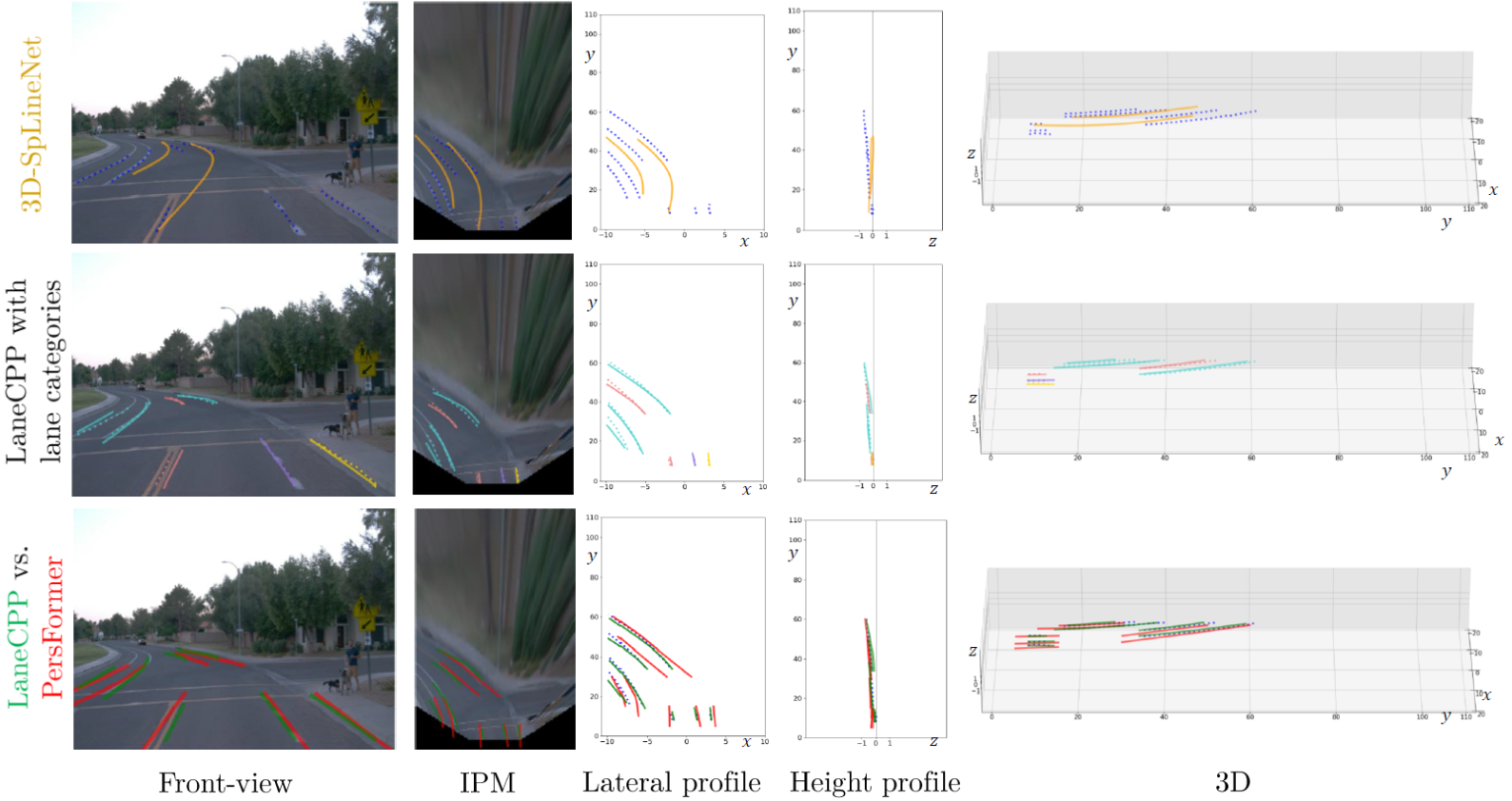}
		\caption{}
		\label{fig:supp-qual-ol-1c}
	\end{subfigure}
	\begin{subfigure}{.97\textwidth}
		\centering
		\includegraphics[width=.99\linewidth]{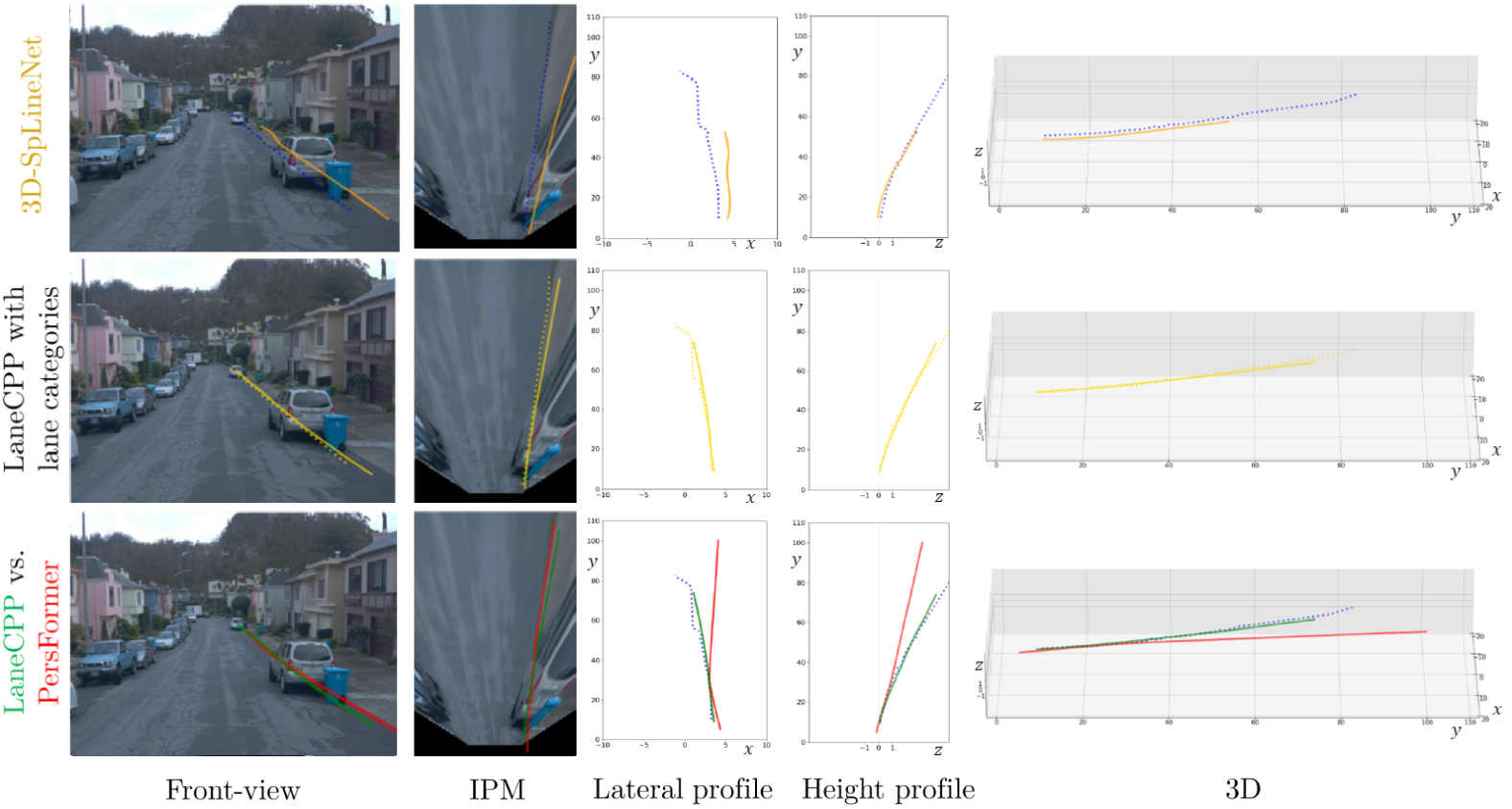}
		\caption{}
		\label{fig:supp-qual-ol-2a}
	\end{subfigure}
	\caption{Additional qualitative evaluation on OpenLane \cite{chen2022persformer} test set (2/3). Top row shows \textcolor{orange}{3D-SpLineNet} baseline compared to \textcolor{blue}{ground truth}. Middle row shows LaneCPP with different lane categories illustrated in different colors and ground truth in dashed lines. Bottom row shows direct comparison of \textcolor{darkgreen}{LaneCPP} and \textcolor{red}{PersFormer*}.}
	\label{fig:supp-qual-ol-2}
\end{figure*}

\begin{figure*}\ContinuedFloat
	\centering
	\begin{subfigure}{.97\textwidth}
		\centering
		\includegraphics[width=.99\linewidth]{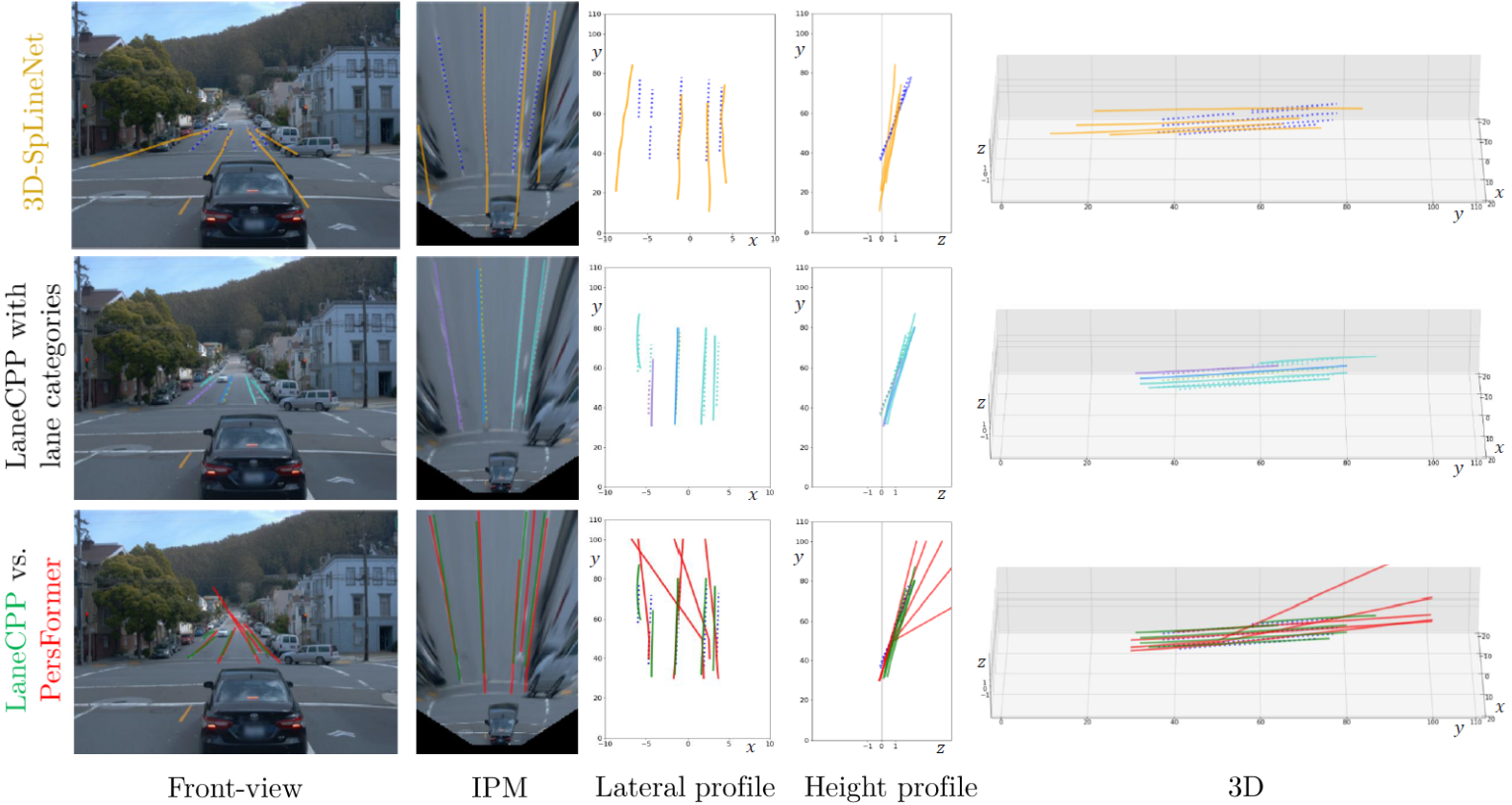}
		\caption{}
		\label{fig:supp-qual-ol-2b}
	\end{subfigure}
	\begin{subfigure}{.97\textwidth}
		\centering
		\includegraphics[width=.99\linewidth]{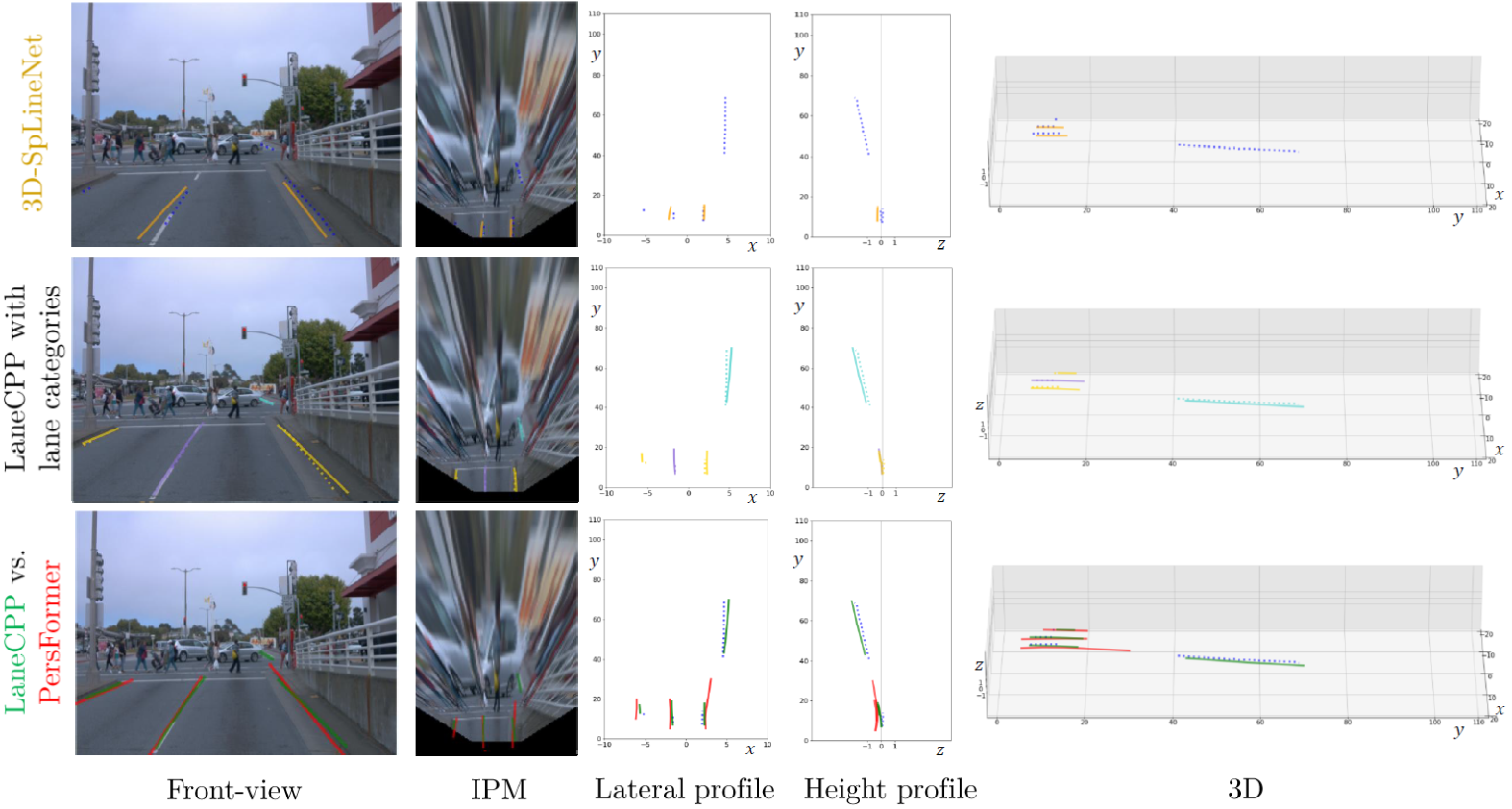}
		\caption{}
		\label{fig:supp-qual-ol-2c}
	\end{subfigure}
	\caption{Additional qualitative evaluation on OpenLane \cite{chen2022persformer} test set (3/3). Top row shows \textcolor{orange}{3D-SpLineNet} baseline compared to \textcolor{blue}{ground truth}. Middle row shows LaneCPP with different lane categories illustrated in different colors and ground truth in dashed lines. Bottom row shows direct comparison of \textcolor{darkgreen}{LaneCPP} and \textcolor{red}{PersFormer*}.}
	\label{fig:supp-qual-ol-3}
\end{figure*}

\begin{figure*}
	\centering
	\begin{subfigure}{.93\textwidth}
		\centering
		\includegraphics[width=.99\linewidth]{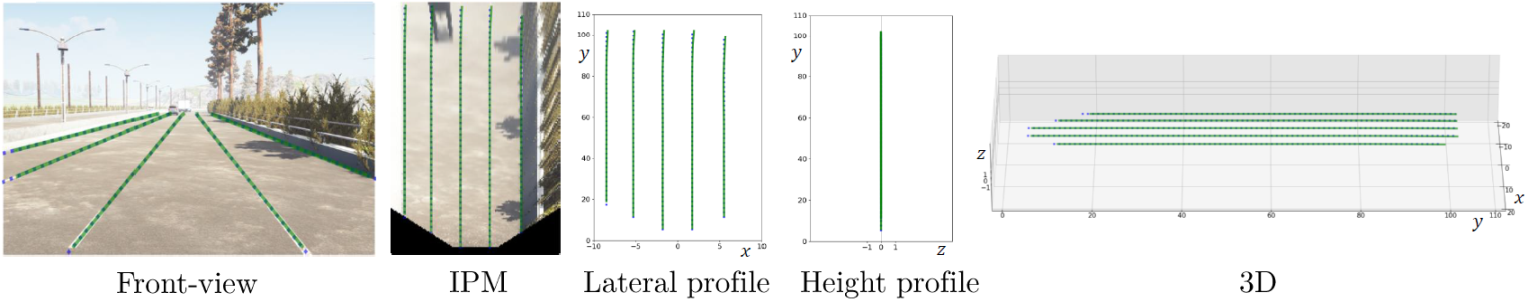}
		\caption{}
		\label{fig:supp-qual-sim-1}
	\end{subfigure}
	\begin{subfigure}{.93\textwidth}
		\centering
		\includegraphics[width=.99\linewidth]{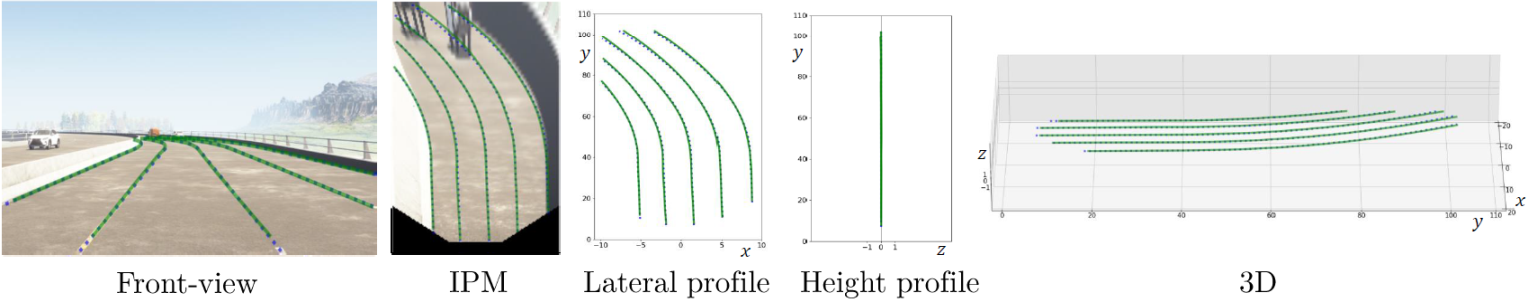}
		\caption{}
		\label{fig:supp-qual-sim-2}
	\end{subfigure}
	\begin{subfigure}{.93\textwidth}
		\centering
		\includegraphics[width=.99\linewidth]{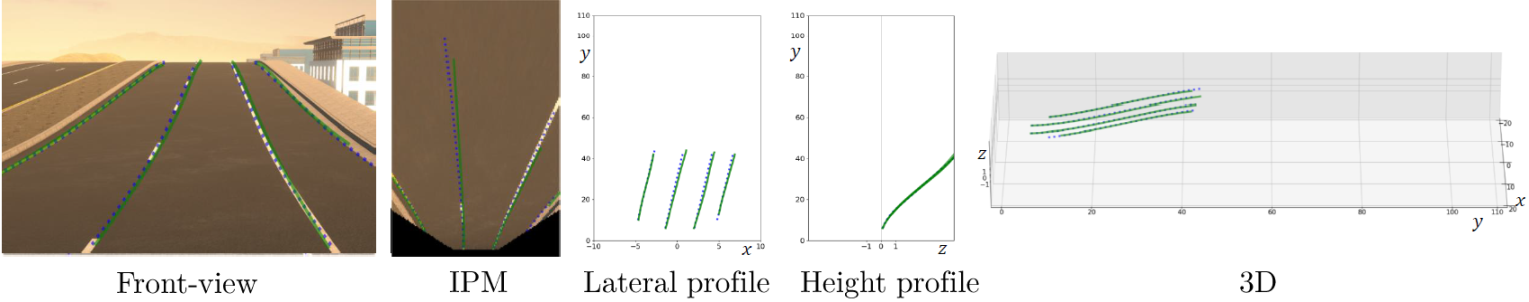}
		\caption{}
		\label{fig:supp-qual-sim-3}
	\end{subfigure}
	\begin{subfigure}{.93\textwidth}
		\centering
		\includegraphics[width=.99\linewidth]{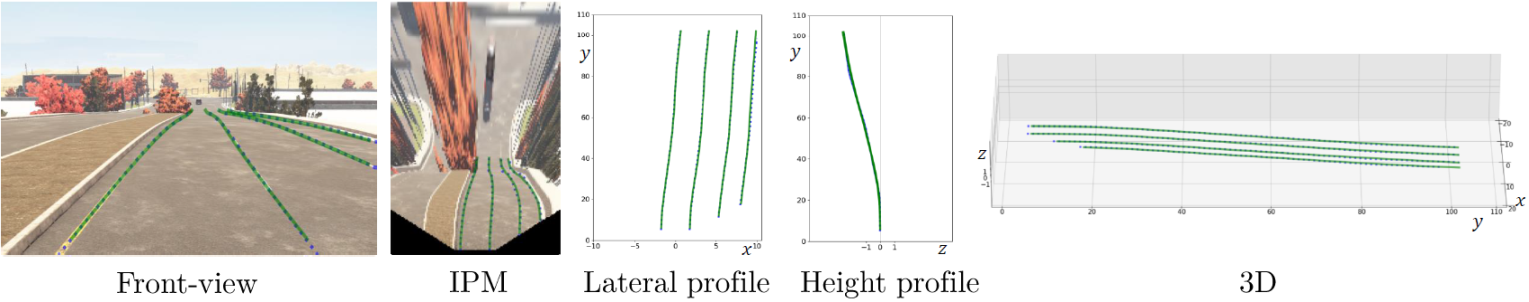}
		\caption{}
		\label{fig:supp-qual-sim-4}
	\end{subfigure}
	\begin{subfigure}{.93\textwidth}
		\centering
		\includegraphics[width=.99\linewidth]{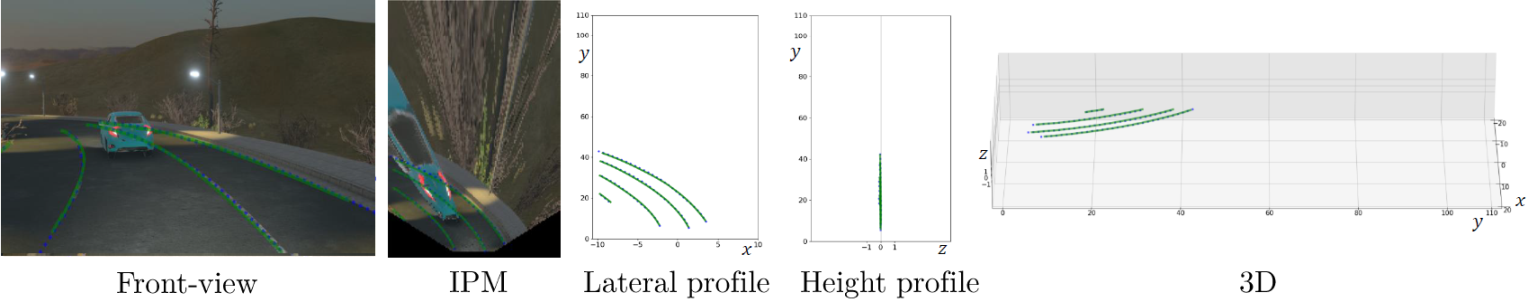}
		\caption{}
		\label{fig:supp-qual-sim-5}
	\end{subfigure}
	\begin{subfigure}{.93\textwidth}
		\centering
		\includegraphics[width=.99\linewidth]{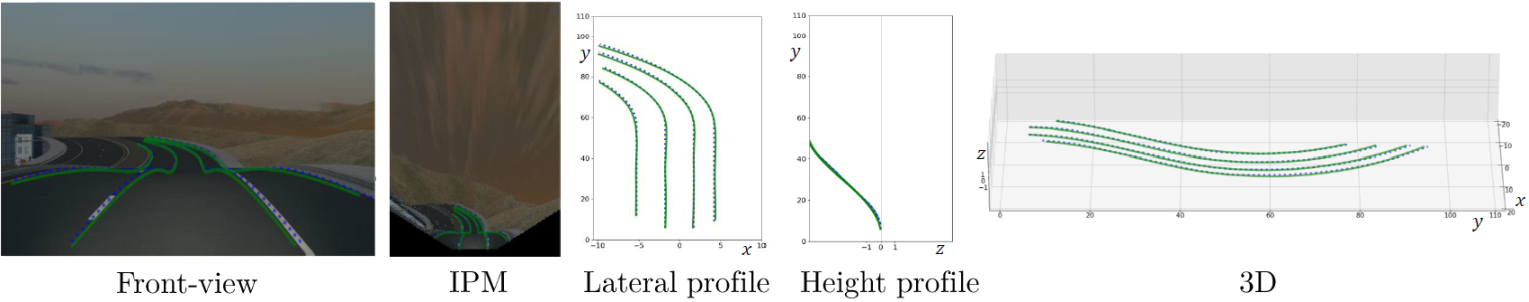}
		\caption{}
		\label{fig:supp-qual-sim-6}
	\end{subfigure}

	\caption{Qualitative evaluation on Apollo 3D Synthetic \cite{genlanenet}. \textcolor{darkgreen}{Our method} is compared to the \textcolor{blue}{ground truth} visualized dashed.}
	\label{fig:supp-qual-sim}
\end{figure*}

We further demonstrate the results of our model on Apollo 3D Synthetic illustrated in \figref{fig:supp-qual-sim}. As shown, our model achieves accurate detection results in simple scenarios from the Balanced Scenes test set (\figref{fig:supp-qual-sim-1} - \figref{fig:supp-qual-sim-2}), in more challenging up- and down-hill scenarios from the Rare Scenes test set (\figref{fig:supp-qual-sim-3} - \figref{fig:supp-qual-sim-4}) as well as in case of visual variations (\figref{fig:supp-qual-sim-5} - \figref{fig:supp-qual-sim-6}). A very challenging scene is shown in \figref{fig:supp-qual-sim-6}, where our model manages to capture the overall line structure well but still could be improved slightly with respect to close-range $x$-errors.

\end{document}